\documentclass[preprint,12pt,authoryear]{elsarticle}

\usepackage{graphicx}
\usepackage{caption}
\usepackage{subcaption}

\usepackage{float}

\usepackage{algorithm}
\usepackage{algpseudocode}

\usepackage[normalem]{ulem}

\usepackage{amsthm,amsmath,amssymb,url,longtable,booktabs,multicol,multirow,lineno}

\usepackage[multiple]{footmisc}
\usepackage{tikz}
\usepackage{tikzscale}
\usepackage{xspace}
\usepackage{pgfplots}
\usepgfplotslibrary{groupplots}
\pgfplotsset{compat=1.16}

\usetikzlibrary{fpu}

\usetikzlibrary{external}
\newcommand{\hamcyc}{\textsc{Hamiltonian Cycle} }
\newcommand{\ts}{\textsc{Travelling Salesman} }
\newcommand{\hcp}{$\mathsf{HCP}$ }
\newcommand{\hcpp}{$\mathsf{HCP}$}
\newcommand{\hc}{\textsc{HC} }
\newcommand{\hcc}{\textsc{HC}}
\newcommand{\tsp}{$\mathsf{TSP}$ }
\newcommand{\tspp}{$\mathsf{TSP}$}
\newcommand{\atsp}{$\mathsf{ATSP}$ }
\newcommand{\atspp}{$\mathsf{ATSP}$}

\newcommand{\npp}{$\mathsf{NP}$}
\newcommand{\fhcpsc}{$\mathsf{FHCPSC}$}
\newcommand{\fhcpscc}{$\mathsf{FHCPSC}$ }
\newcommand{\lkh}{$\mathsf{LKH}$ }
\newcommand{\lkhh}{$\mathsf{LKH}$}

\newcommand{\ma}{$\mathsf{MA}$ }
\newcommand{\maa}{$\mathsf{MA}$}
\newcommand{\rai}{$\mathsf{RAI}$ }

\newcommand{\tc}{$\mathsf{TC}$ }
\newcommand{\tcc}{$\mathsf{TC}$}

\newcommand{\sr}{$\mathsf{SR}$ }
\newcommand{\srr}{$\mathsf{SR}$}

\newtheorem{theorem}{Theorem}
\theoremstyle{definition}

\usepackage{breakcites} 

\usepackage{hyperref}

\journal{Artificial Intelligence}
\newtheorem{definition}{Definition}[section]
\begin{document}

\begin{frontmatter}

\title{A Memetic Algorithm To Find a Hamiltonian Cycle in a Hamiltonian Graph
}

\author{Sarwan Ali$^1$}
\author{Pablo Moscato$^2$
\\
$^1$Department of Computer Science, Georgia State University, Atlanta, Georgia, USA
\\
$^2$College of Engineering, Science and Environment, The University of Newcastle, Callaghan, NSW 2308, Australia
\\
sali85@student.gsu.edu, Pablo.Moscato@newcastle.edu.au
}





\begin{abstract}
We present a memetic algorithm (\maa) approach for finding a Hamiltonian cycle in a Hamiltonian graph. The \ma is based on a proven approach to the Asymmetric Travelling Salesman Problem (\atspp) that, in this contribution, is boosted by the introduction of more powerful local searches. Our approach also introduces a novel technique that sparsifies the input graph under consideration for Hamiltonicity and dynamically augments it during the search. Such a combined heuristic approach helps to prove Hamiltonicity by finding a Hamiltonian cycle in less time. In addition, we also employ a recently introduced polynomial-time reduction from the \hamcyc to the Symmetric \tsp, which is based on computing the transitive closure of the graph. 
Although our approach is a metaheuristic, i.e., it does not give a theoretical guarantee for finding a Hamiltonian cycle, we have observed that the method is successful in practice in verifying the Hamiltonicity of a larger number of instances from the \textit{Flinder University Hamiltonian Cycle Problem Challenge Set} (\fhcpsc), even for the graphs that have large treewidth. The experiments on the \fhcpscc instances and a computational comparison with five recent state-of-the-art baseline approaches show that the proposed method outperforms those for the majority of the instances in the \fhcpsc. 

\end{abstract}





\end{frontmatter}

\section{Introduction}\label{intro}
Given as input a graph $G(V,E)$ where $V$ is the set of vertices and $E$ is the set of edges, the task of deciding if $G$ has a Hamiltonian cycle (i.e., a cycle of length $\vert V \vert$) in $G$ is \npp-complete. The \hamcyc problem (\hcpp) is indeed one of the most conspicuous members of the \npp-complete class~\cite{karp1972reducibility,baniasadi2014deterministic}. In this communication, we address the problem of finding such a cycle when we know that there is at least one. We are not solving the decision version in this contribution, but we present a heuristic to find such a Hamiltonian cycle in $G$ (\hc) when we know that $G$ is Hamiltonian.  

A lot of effort has been made over the years to solve the \hcpp. While this problem is presented in lectures to nearly all computer science undergraduates around the world, and it is well known, the \hcpp still attracts considerable attention, particularly since it is a challenge for some types of instances that are very large and have some structural parameters that make them hard in practice for current algorithms and heuristics.

The approach of finding a \hc  in a graph by reducing a generic instance of the \hcp to an instance of the \textit{Travelling Salesman Problem} (\tspp) continues to be in practice~\cite{baniasadi2014deterministic} with
new variants being recently proposed~\cite{DBLP:conf/ssci/MathiesonM20}. 
Towards this end, we need to find the tour of minimum length (where each edge has an associated weight, and the length of the tour is the sum of the weights of edges comprising that tour). There are many exact algorithms and heuristics proposed in the literature to solve the \tsp, like~\cite{Concorde} and~\cite{HELSGAUN2000106}. For further detail, readers are referred to~\cite{gutin2006traveling},~\cite{lawler1985traveling},~\cite{svensson2020constant}, and~\cite{applegate2006traveling}. 
Although some of these methods perform reasonably well towards solving the \tsp in a practical sense, they failed to deliver, for instance, the \hcp with particular structures and large sizes~\cite{DBLP:conf/evoW/AhammedM11, DBLP:journals/mpc/HougardyZ21}.

\subsection{The Flinders University \hamcyc Problem Challenge Set (\fhcpsc)}
\begin{figure}[h!]
\minipage{0.48\textwidth}
  \includegraphics[height = 7cm,width=\linewidth] {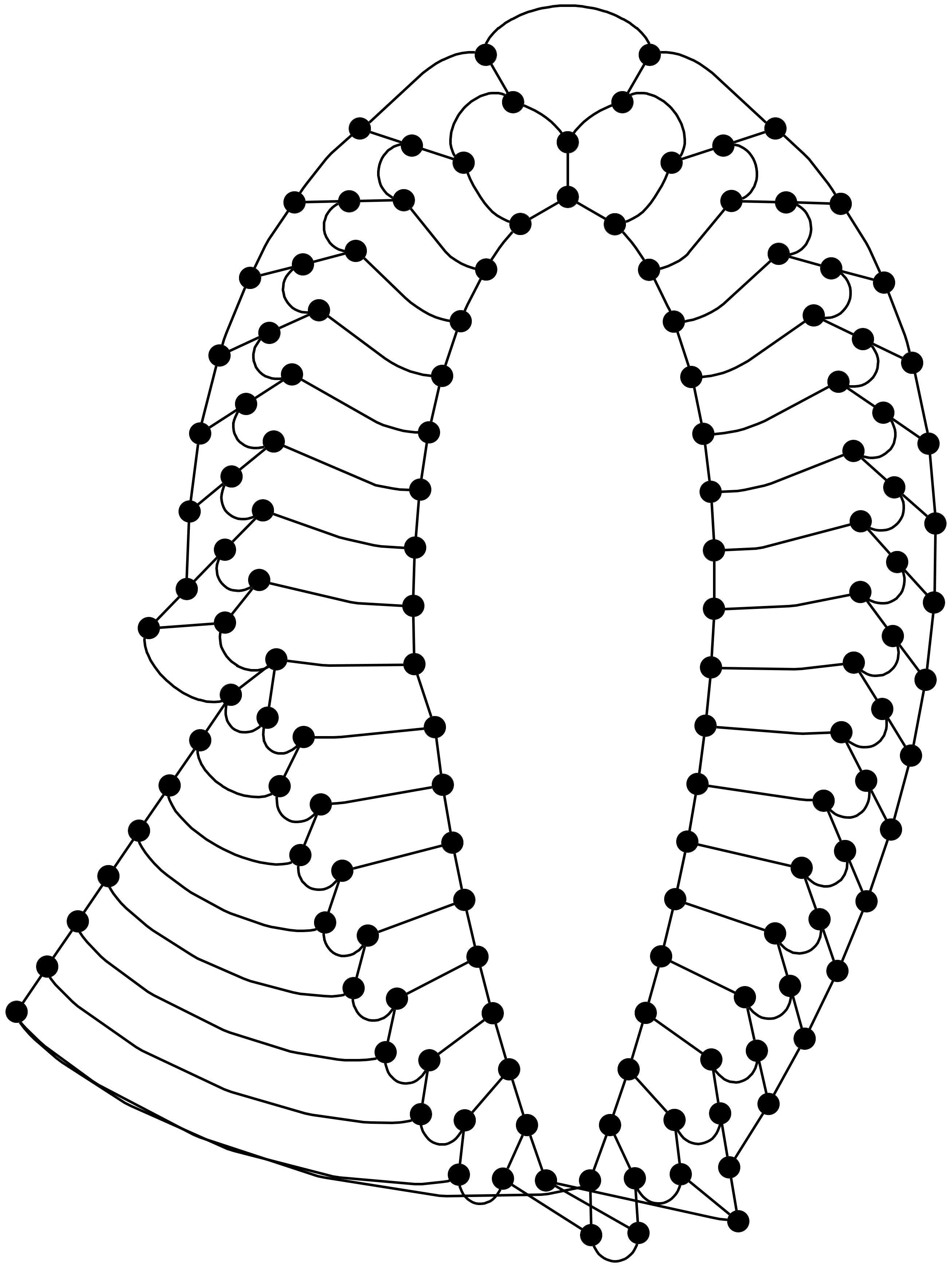}
  \caption*{(a) Instance 14}
\endminipage\hfill
\minipage{0.48\textwidth}
  \includegraphics[height = 7cm,width=\linewidth] {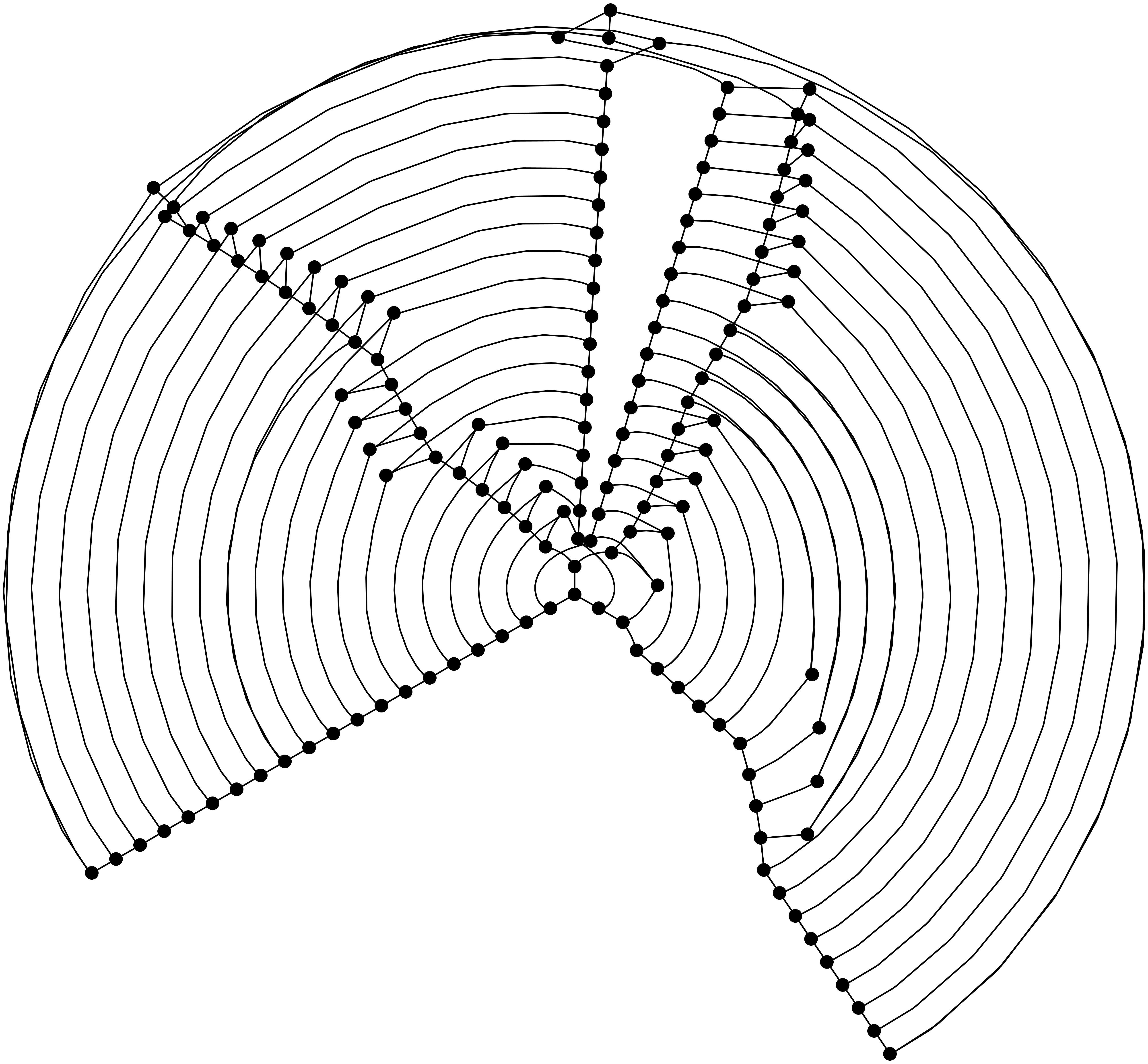}
  \caption*{(b) Instance 16}
\endminipage\hfill
\caption{Different structures of graphs (using a radial layout of the graph visualization software suite $\texttt{yEd}$) in the Flinders University \hamcyc Problem Challenge Set. The Treewidth of both instances is $9$.}
\label{fig_graph_structures}
\end{figure}

A challenging dataset of Hamiltonian graph instances was introduced in 2015~\cite{Haythorpe2015}. These are known as the \textit{Flinders University \hamcyc Problem Challenge Set} (\fhcpsc). The whole database consists of $1001$ instances of a wide variety of types, all having in common that they are known to have at least one Hamiltonian cycle. All of the instances are designed to be difficult to solve using standard \hcp heuristics. These graphs range in size from $66$ vertices up to $9528$ vertices, and they have an average size of just over $3000$ vertices (see Figure~\ref{fig_graph_structures} to analyze two different structures of graphs in \fhcpsc, which is drawn using a \textit{radial layout} algorithm of the graph visualization software suite $\texttt{yEd}$\footnote{\url{https://www.yworks.com/products/yed}}). 
Several efforts have been made in the literature to solve as many instances of the \fhcpscc as possible. To the best of our knowledge, no algorithm or heuristic has solved all of these instances so far.

\subsection{Recent studies}
Previous studies on \hcp have been inspired by real-life applications~\cite{rahman2005Hamiltonian,applegate2006traveling}.
There are many applications of \hc in a wide range of fields such as electronic circuit design, DNA sequencing, operations research, computer graphics, and mapping genomes~\cite{ravikumar1992parallel,dogrusoz1995hamiltonian,grebinski1998reconstructing}. 
Besides, schools and colleges use Hamiltonian cycles to plan the best route to pick up students from different locations.

The \hcp has also been used by researchers for computational challenges, leading to experimentation and open data sharing. The overarching aim of the researchers is to benchmark the performance of existing exact algorithms and heuristics.  Exact algorithms (e.g., branch-and-bound, integer programming formulations, cutting planes, and Lagrangian dual~\cite{cinar2020discrete}) can solve the \tsp to optimality, which is obviously a challenge for larger instances~\cite{applegate2006traveling}. Heuristics (and meta-heuristics such as neural networks, tabu search, cuckoo search, artificial immune systems, and imperialist competitive algorithm~\cite{cinar2020discrete}, etc.) and, in particular memetic algorithms, use problem-specific knowledge to find optimal/near-optimal solution of the \tspp. Although heuristics do not guarantee solving the \tsp optimally, they usually scale well on the larger instances (for optimally finding the tour). However, heuristics are usually problem-specific, which means that different heuristics have to be designed for different types of problems and for different instances of the same problem. A comprehensive description of alternative related methods is included in Section~\ref{sec_related_work}.

Designing a heuristic method, which can generalize on different types of problems, is very challenging and draws the attention of a lot of researchers~\cite{burke2emerging}. Towards this end, we propose a heuristic in this paper to solve the \hcpp, which is intended to be largely independent of the structure of the graph.

\subsection{Our contributions}

In this paper, we present a memetic algorithm that employs local search, dynamic programming approaches, and a sparsification strategy. 
A memetic algorithm is a population-based approach for problem-solving based on a set of competing and cooperating agents. It has successfully been used in the past to solve difficult \npp-complete and \npp-hard 
problems such as vertex cover~\cite{wu2018game}, the asymmetric \tsp ~\cite{DBLP:journals/heuristics/BuriolFM04}, Hamiltonian cycle problem~\cite{shaikh2012solving}, quadratic assignment problem~\cite{merz1999comparison}, and time series compression~\cite{friedrich2020memetic} just to mention a few applications.
The bibliographic search engine of \textit{Web of Science} retrieves nearly $4000$ manuscripts on this meta-heuristic approach, which is well-known and consolidated right now.
We refer to the work done by~\cite{DBLP:books/sp/19/MoscatoM19} for a recent survey of nearly $600$ applications in data science and analytics. 

We consider \ma as a leading case of a more general algorithmic design pattern template for meta-heuristics approaches, based on over-constraining the original problem instance (by introducing dynamically changing new restrictions), together with a dramatic reduction of the configuration search space (here due to sparsification of the original graph) and the iterative use of powerful local search solvers. 

Therefore, the main contributions of this work are:
\begin{itemize}
    \item Our proposed heuristic approach is simple to adapt from an existing MA for the ATSP/TSP, yet it is able to solve $655$ instances (out of $1001$) of \fhcpscc challenge set. We were able to solve more instances than the proposal, which got the second position in the \fhcpscc challenge (that solved $614$ instances).
    In passing, we note that the winner of the challenge (discussed in Section~\ref{sec_most_solved_instances}) used a ``human-machine approach'' that has not been completely described in algorithmic terms, and to the best of our knowledge, no computer implementation currently exists or is being planned. 
    \item Our algorithmic approach is quite general and can be applied to any type of Hamiltonian graph, regardless of its structure.
    \item We present computational results that improve state-of-the-art approaches for the \hcp in terms of runtime for the majority of the instances. We compare our proposed algorithm with five baseline methods, and we show improvements in terms of runtime.
    \item We were also able to solve complex instances (e.g., large treewidth) that were not previously solved by any other baseline method (a formal definition of the concepts of treewidth and tree decomposition of a graph are given in Section~\ref{sec_preliminaries}).
\end{itemize}

\subsection{Organization of this manuscript}

The rest of the manuscript is organized as follows. 
In Section~\ref{sec_preliminaries}, we introduce some basic terms and notations related to graphs in general, which have been used throughout the manuscript. In Section~\ref{algorithmic_component}, we define different techniques, which we are using in our proposed method. Section~\ref{sec_propose_approach} discusses our proposed memetic algorithm in detail. The dataset description, baseline methods, and implementation details of our algorithm are given in Section~\ref{sec_experimental_setup}. We present our results in Section~\ref{results_and_discussion}. The conclusion of our work is given in Section~\ref{section_conclusion}. Lastly, we discuss some aspects of related work in Section~\ref{sec_related_work}.

\section{Basic definitions}\label{sec_preliminaries}
Given an input graph $G$,  we use $V(G)$ and $E(G)$ to denote the vertex and edge sets of $G$, respectively. All graphs are undirected, with neither self-loops nor parallel edges.
Our goal is to find a cycle (Hamiltonian~\footnote{
The name came from Sir William Hamilton’s investigations of such cycles on the dodecahedron graph around $1856$} cycle) in $G$ of length $n=|V|$. We start by defining a few terms, which will be consistently used throughout the writeup.

\begin{definition}[Tree Decomposition] -
For a graph $G = (V,E)$, a tree decomposition (TD) of $G$ consists of a tree $\mathcal{T}$ and a subset $V_{t} \subseteq V$ (often called a \textit{bag}), where $V_{t}: t \in \mathcal{T}$. The tree $\mathcal{T}$ and the bag must satisfy the following properties: (i) every node $v \in V(G)$ must belong to at least one bag $V_{t}$ (node convergence), (ii) for every edge $e \in E(G)$, there is some bag containing both ends of $e$ (edge convergence), and (iii) let $A, B, C \in \mathcal{T}$, such that $B$ lies on the path from $A$ to $C$. If a node $v \in V(G)$ belongs to both $A$ and $C$, then it must also belong to $B$ (coherence). 
\end{definition}
\begin{definition}[Width of a Tree Decomposition] - 
The \textit{width of a tree decomposition} given a tree $\mathcal{T}$ and $V_{t}$ (where $V_{t} \subseteq V$ and $V_t : t \in \mathcal{T}$ ) equals to $max_{t \in V_{\mathcal{T}}} \vert V_{t} \vert - 1$ (where $\vert V_{t} \vert$ is the size of the largest subset/bag of vertices $V$). 
\end{definition}
\begin{definition}[Treewidth] - 
Given all possible TDs of a graph $G$,
the \textit{treewidth} of $G$ is the minimum possible width of a TD among all TDs.
\end{definition}

In order to explain the sparsification heuristic, we found it useful to refer to the following two concepts. 

\begin{definition}[Conflicting Edge of a graph $G(V,E)$ given $G'(V,E')$] -
\label{def_conflicting}
Given two graphs $G$ and $G'$, where $V'(G') = V(G)$ and 
$E'(G') \neq E(G)$, a conflicting edge is any edge $e \in E'$ for which $e \notin E$.
\end{definition}

Then the set of all conflicting edges is $\Delta(E',E) = E'-E$, so in our sparsification heuristic, we will try to minimize both $\Delta E$ and $|E'|$ while, hopefully, still maintain the Hamiltonicity of $G'$.

\begin{definition}[Surplus edge of a graph $G(V,E)$ given a Hamiltonian cycle $H$ of $G$] - \label{def_surplus}
Given a graph $G$, $E(H)$ is the set of edges in the Hamiltonian cycle in $G$ (if present). 
We will say that an edge $e \in E$, given $H$, is called a surplus edge of $G$ given $H$
iff $e \in E(G) - E(H)$.
\end{definition}

The goal of sparsification is then to reduce the number of surplus edges ($\vert E(G) - E(H) \vert$) in (by creating a sparse version) so that the Hamiltonian cycle can be found in less time using our \maa.

\section{Algorithmic components of the strategy}\label{algorithmic_component}
Although the terminologies and definitions discussed below are well-known for researchers in routing problems and graph optimization, we judge that a short introduction to the main algorithmic components is required for completeness. We start with the main solvers and definitions that are still required. 
\begin{definition}[\textsc{Weighted Hamiltonian Cycle}] - 
Given $G = (V,E,W)$, a simple undirected graph with non-negative integer weights, and an integer $L$, the \textsc{Weighted Hamiltonian Cycle Problem} requires us to decide if $G$ contains a Hamiltonian cycle $C \subseteq E $ of length (total sum of weights) $l(C) \leq L$ or if it does not. 
\end{definition}

When the graph is complete and undirected, we can think of the \textsc{Weighted Hamiltonian Cycle Problem} as the well-known symmetric \tspp. 
Several efforts have been made in the literature to solve the optimization version of the \tspp, i.e., the task of finding the Hamiltonian cycle of the shortest total length in a complete weighted graph. A few \tsp solvers worth of note exist. Some methods are complete and give an optimal solution, for instance Concorde~\cite{Concorde}; others are based on powerful local search techniques (e.g. Lin Kernighan Heuristic (\lkhh)~\cite{lin1973effective,HELSGAUN2000106}) or population-based methods such as memetic algorithms~\cite{moscato1992memetic, 10.5555/329055.329079,DBLP:journals/compsys/MerzF02}.
One common approach to deal with the \hcp is to apply a polynomial-time reduction that transforms a generic instance of the \hcp into a particular instance of the \tspp, and then apply powerful \tsp solvers to find the minimum length tour. We discuss two reduction methods that have been proposed in the literature to transform the \hcp into \tspp. 

\begin{figure}[h!]
	\centering
	\includegraphics[scale=0.4] {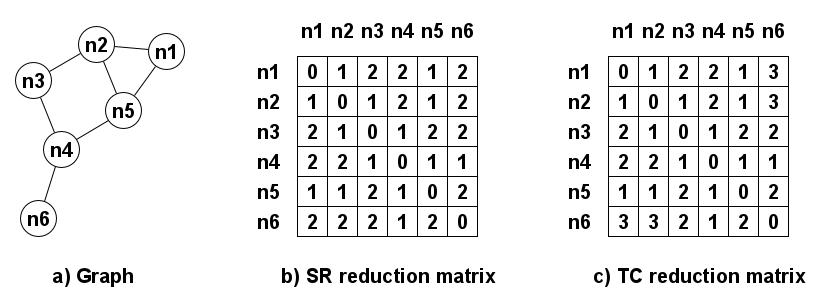}
	\caption{Process of generating \sr and \tc reduction matrix. The sub-figure (a) is an illustrative sample graph for which \sr reduction matrix is generated in the sub-figure (b), and \tc reduction matrix is generated in the sub-figure (c).}
\label{fig_sr_reduction}
\end{figure}

\begin{definition}[Standard Reduction (\srr)] - 
This method is the traditional Karp-reduction taught in many undergraduate courses around the world and described in textbooks such as~\cite{garey1979computers}. An $n \times n$ integer matrix $M$ is generated (where $n = \vert V \vert$). 
Given $G = (V,E)$ for each edge $e \in E$ between two vertices $a$ and $b$ in the graph, 
$M_{a,b}=M_{b,a}=1$ and all other coefficients are set to the value of $2$ (see Figure~\ref{fig_sr_reduction} (b)). 
The distance matrix $M$ is now an instance of the \tspp. An exact solver (such as Concorde) will stop after finding a tour of length $n$, thus indicating that $G$ is Hamiltonian.
\end{definition}

\begin{definition}[Transitive Closure (\tcc) Reduction] - 
This new method has been recently proposed in~\cite{DBLP:conf/ssci/MathiesonM20}. 
The assignments of coefficients of $M$ valued at `1' are the same as in the case of the Standard Reduction. For each pair of vertices $a$ and $b$ which are not connected by an edge in $E(G)$, $M_{a,b}$ and $M_{b,a}$ will be assigned an integer value equal to the distance between $a$ and $b$ in $G$ (see Figure~\ref{fig_sr_reduction} (c)).
\end{definition}


The main heuristic idea behind using the matrix provided by the \tc reduction instead of the one provided by the \sr approach is to create a useful fitness landscape for the memetic algorithm~\cite{Mendes_fitnesslandscapes,Moscato2019,DBLP:conf/cec/NeriZ20}
since it has been postulated since their introduction that the correlation of local minima is an important part of the design of suitable configuration space for population-based search methods for optimization~\cite{DBLP:journals/anor/Moscato93,DBLP:conf/gecco/SapinJS16}. 
The authors of~\cite{DBLP:conf/ssci/MathiesonM20} argue that, although from the theoretical perspective, there is no difference between \tc and \sr to prove \npp-completeness, choosing the \tc reduction can produce a more correlated fitness landscape for the \ma than the standard approach. Therefore, in our experiments, we are using \tc reduction only. For other alternative reduction methods and results, we refer to~\cite{DBLP:conf/ssci/MathiesonM20}. 

\section{Proposed Approach}\label{sec_propose_approach}
We use a highly effective memetic algorithm ($\mathsf{MA}$) that closely follows a previously proposed \ma for the \textsc{Asymmetric TSP} in
\cite{DBLP:journals/heuristics/BuriolFM04}. 
It is based on a tree-structured population
of three levels. Each agent in the population has two solutions, called \textit{Current} and \textit{Pocket}. 
The \textit{Current} solution represents the agent's present location in the solution space and undergoes local search operations to explore the nearby region. This exploration allows the algorithm to discover potentially improved solutions. On the other hand, the \textit{Pocket} solution serves as a memory mechanism, preserving the best solution encountered by the agent throughout its exploration. 
While the \textit{Current} solution focuses on exploring the local neighborhood, the \textit{Pocket} solution retains the best-known solution to prevent its loss during exploration. This dual-solution strategy strikes a balance between exploration and exploitation, facilitating the algorithm in navigating the solution space effectively, avoiding premature convergence to suboptimal solutions, and promoting the discovery of high-quality solutions.
This approach has been used for many other problems, such as symmetric TSP~\cite{Moscato_Tinetti_92}, 
protein structure prediction~\cite{7836019,INOSTROZAPONTA2020101087},
multivariate regression~\cite{DBLP:conf/cec/SunM19,DBLP:conf/cec/MoscatoSH20},
the quadratic assignment problem\cite{DBLP:conf/cec/HarrisBIM15},
the lot sizing problem~\cite{BERRETTA200467},
number partitioning~\cite{Berretta2004},
and many others. While we follow the approach of~\cite{DBLP:journals/heuristics/BuriolFM04} in this case, we use more powerful local search techniques to boost its performance.  

One of the differentiating characteristics of this \ma is that we are using a more powerful local search heuristic \lkh (such as in~\cite{DBLP:journals/compsys/MerzF02}), but we also maintain a local search heuristic originally designed for the \atsp called \textit{Recursive Arc Insertion} (RAI)~\cite{DBLP:journals/heuristics/BuriolFM04}. 
The \ma will be using the matrix obtained by the application of the \tc reduction.  
The other important characteristic is that we now apply an efficient sparsification technique on the distance matrix to reduce the search space (hence, we heuristically aim at reducing the treewidth of the graph) 
thus facilitating the work of the \maa. 
This means that the \ma has the distance matrix that is changed during the run as a consequence of the sparsification heuristic.
The fourth important different thing to notice is that our \ma also uses another method, called the \textit{Na\"{\i}ve Algorithm} (see Section~\ref{subsection_naive_algorithm}) proposed by~\cite{bodlaender2015deterministic}  to further enhance its efficiency. The combination of local search methods, sparsification of instances, and the use of an ad hoc approach together help us enhance the overall quality of our \maa.

Like the one of~\cite{DBLP:journals/heuristics/BuriolFM04}, this \ma also uses a tree-structured population of $13$ agents, but now finding \hc using \rai and \lkh local search heuristics together.
Initially, our algorithm runs the so-called Na\"{\i}ve Algorithm~\cite{bodlaender2015deterministic} to try to find a Hamiltonian cycle. 
Note that the Na\"{\i}ve Algorithm takes uses an edgelist of the original graph $G$ as an input (without applying \tc reduction) and returns the \hc if exist.
If the solution is found (using Na\"{\i}ve Algorithm), the algorithm immediately terminates. If the contrary is true, the subsequent population-based operation of our \ma starts.  The basic structure of our proposed algorithm is given in Section~\ref{sec_memtic_algo_structure}.

\subsection{Na\"{\i}ve Algorithm} \label{subsection_naive_algorithm}
The Na\"{\i}ve Algorithm uses a dynamic programming approach to find the \hc in a given graph~\cite{cygan2015parameterized}. It starts by building multiple partial solutions. To construct the partial solutions in the subgraphs of the original graph, it uses a bottom-up approach to the tree decomposition. 

Consider a separation ($A,B$) in a graph $G$, where $A, B \subseteq V (G) $, $A \cup B = V (G)$, and each edge of $G$ belongs to exactly one of the sub-graphs $A$ or $B$. 
Suppose that there is no edge between $A \setminus B$ and $B \setminus A$. 
The partial solution for separation $(A,B)$ in $G[A]$ would be a set of vertex-disjoint paths $\mathcal{P}$ that all have endpoints in $S$ (where $S = V(A) \cap V(B)$). These paths together visit every vertex of $A \setminus B$. 
To complete the partial solution $\mathcal{P}$ to a Hamiltonian cycle $H$ in $G$, the algorithm keeps track of the vertices of $S$, which are visited by $\mathcal{P}$. It also keeps track of the endpoints of paths (and how these paths pair up with endpoints) in $\mathcal{P}$. Given a tree decomposition (of width $t$) of an input graph $G$, the Na\"{\i}ve Algorithm finds the Hamiltonian cycle in $2^{O(t \ log \ t)}n$ time, where $n$ is the total number of vertices.
For more detail related to treewidth and dynamic programming, readers are referred to~\cite{cygan2015parameterized}. 

\subsection{Our Memetic Approach} \label{sec_memtic_algo_structure}
\begin{figure}[h!]
	\centering
	\includegraphics[scale=0.35] {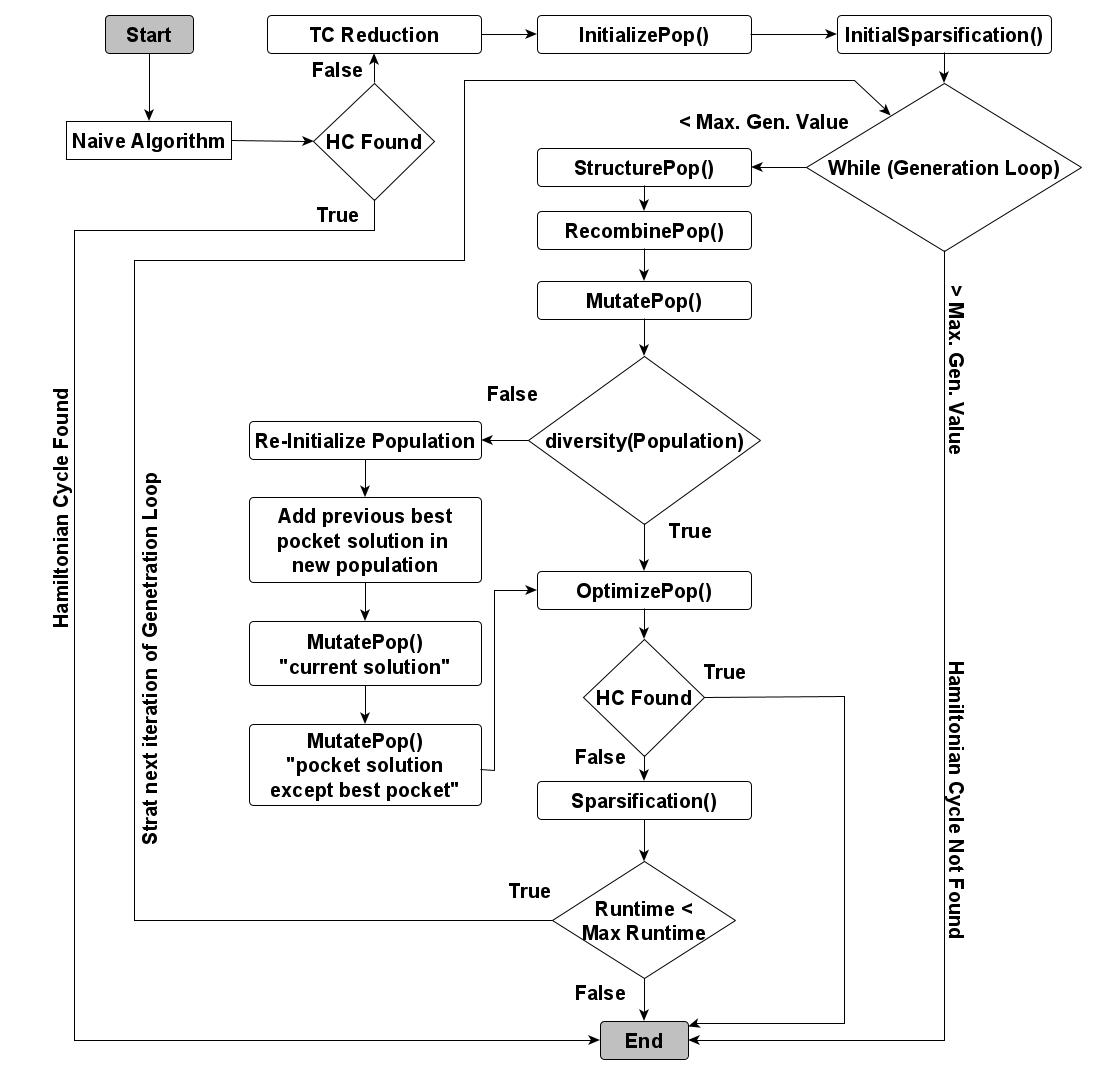}
	\caption{Flow Chart of our proposed approach.}
\label{fig_flow_chart}
\end{figure}

The pseudo-code of our proposed \ma is given in Algorithm~\ref{algo_1}.
We first apply the Na\"{\i}ve algorithm on the graph. If the solution is not found, the population search is started by initializing thanks to a nearest neighbor heuristic. Then, we apply sparsification on the graph to reduce the search space. After that, we restructure the population based on the solution's cost. After restructuring, recombination and mutation are applied. Then, we apply \lkh and RAI local search to optimize the results. In the end, we apply an augmentation approach to include more edges in the sparse graph.
Figure~\ref{fig_flow_chart} contains the flow chart of our whole proposed approach. We will now describe each procedure of Algorithm~\ref{algo_1} one by one.

\begin{algorithm}[h!]
    \centering
    \caption{Pseudocode of our \ma Algorithm}
    \begin{algorithmic}[1]
    \State apply the Na\"{\i}ve Algorithm on $G$ \Comment{see Section~\ref{subsection_naive_algorithm} for Na\"{\i}ve Algorithm}
    \If{solution not found using Na\"{\i}ve Algorithm}
        \State initializePop() \Comment{based on a nearest neighbor heuristic}
        \State initialSparsification() \Comment{based on the \lkh heuristic}
            \Repeat /* generation loop */ \Comment{see Equation~\ref{eq_max_generations} for max generation value}
            \State structurePop() \Comment{organize the ternary tree population}
            \State recombinePop() \Comment{SAX operator is used}
            \State mutatePop() \Comment{bring diversity in the population}
            \If{diversity(Population) == false} \Comment{When \# of Generations $> 30$}
                \State restartPop() \Comment{restart to avoid population stagnation}
            \EndIf
            \State optimizePop() \Comment{using both \rai and \lkh local search}
            \State sparsification() \Comment{based on the \lkh heuristic}
            \Until {TerminationCondition = satisfied} \Comment{see Section~\ref{sec_stopping_criteria} for termination conditions}
        \EndIf
    \end{algorithmic}
    \label{algo_1}
\end{algorithm}

\subsubsection{Population Initialization}
In our algorithm, the \textit{initializePop()} function is used to initialize the population of agents using the well-known Nearest Neighbor Heuristic (as described in~\cite{DBLP:journals/informs/MoscatoN98}, also see Definition~\ref{nn_definition}) 
with the variant implemented in~\cite{DBLP:journals/heuristics/BuriolFM04}.
Although this initialization quickly yields short tours, none may be optimal, and the next step of population-based search generally starts.
\begin{definition}[Nearest Neighbor Heuristic]\label{nn_definition} - 
Starting from a random city, select the city that is closest (having minimum distance) to the current city. If two unvisited cities have the same distance from the current city, we randomly select one city from those two unvisited cities. Repeat the process until all cities are visited.
\end{definition}

\subsubsection{Initial Sparsification}
In order to find a Hamiltonian cycle in a graph $G$, we can think of a partition
of $E(G)$ in two groups: those that form one (unknown) Hamiltonian cycle (we can call them \textit{Hamiltonian edges}) and the others that are 
not in that Hamiltonian cycle (we call these the \textit{surplus edges}, see Definition~\ref{def_surplus}).
As a heuristic attempt to identify candidates that may belong to the surplus edge set, we first apply the 
\lkh heuristic to the top/leader agent's pocket solution to obtain a \lkh locally optimal tour $T$. 

\begin{figure}[htpb!]
\minipage{0.5\textwidth}
  \includegraphics[scale=0.2]{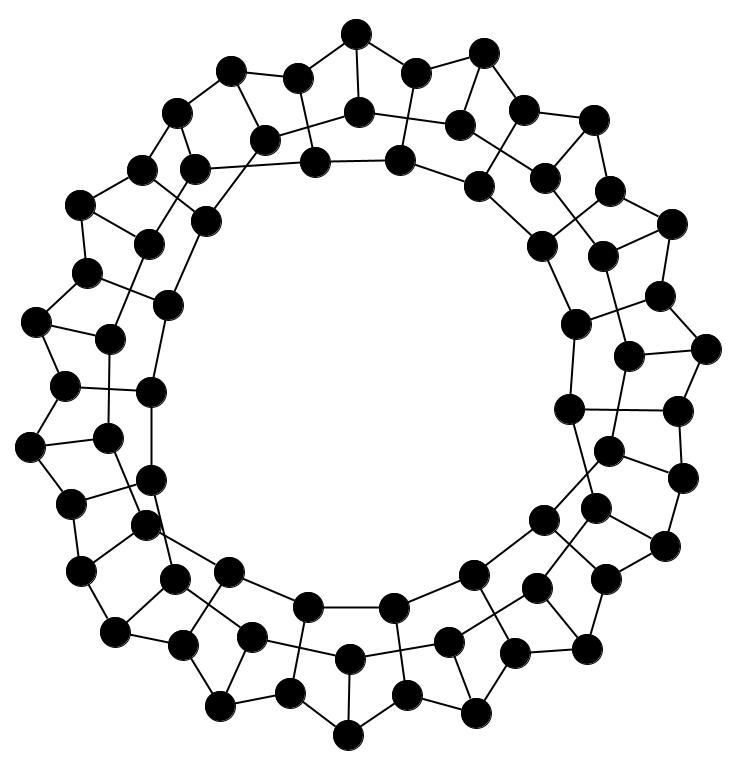}
  \caption*{(a) Original Graph}
\endminipage\hfill
\minipage{0.5\textwidth}
  \includegraphics[scale=0.2]{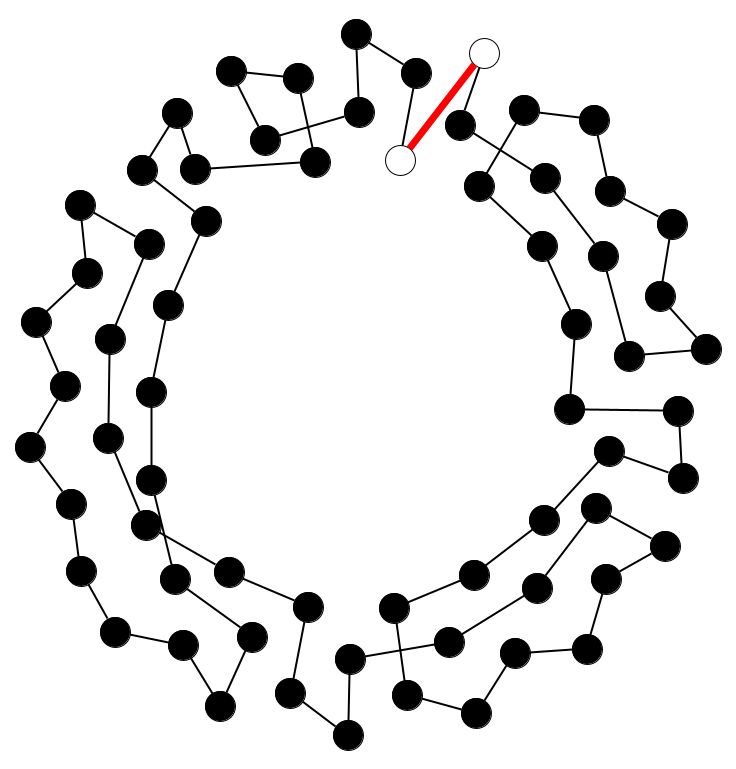}
  \caption*{(b) Sparse Graph after applying \lkh } 
\endminipage
\hfill
\\
\begin{center}
\minipage{0.5\textwidth}%
  \includegraphics[scale=0.2]{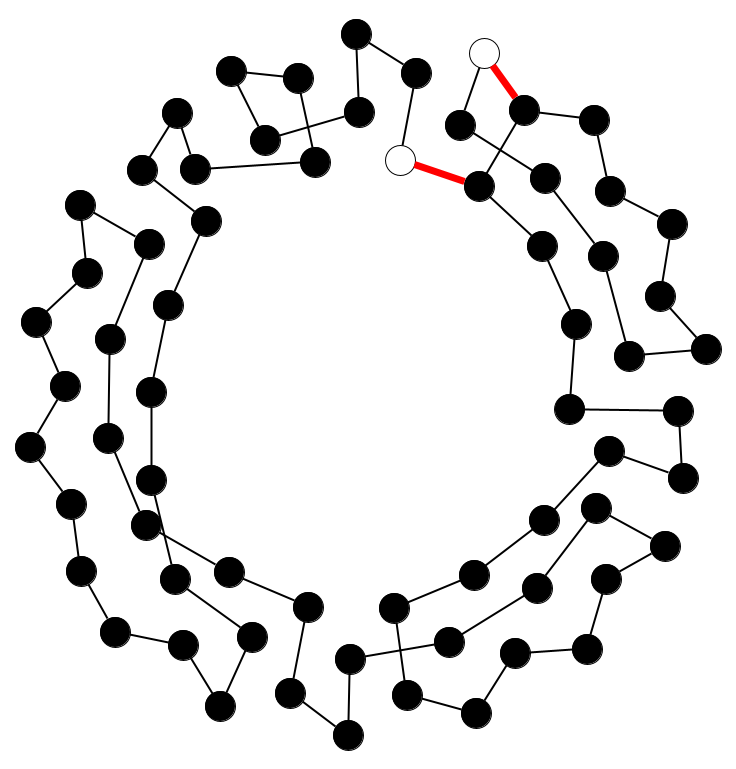}
  \caption*{(c) Sparse Graph after eliminating conflicting edges}
\endminipage
\end{center}
\caption{Steps to make the graph sparse. The sub-figure (a) is the original graph (instance 1). After running \lkh on (a), we get a tour, which is represented in sub-figure (b). Note that there are two vertices with white color in (b), which are the endpoints of the conflicting edge (edge in red color). To deal with the conflicting edge problem, we remove the conflicting edge and add edges in the shortest path (given in red color) between the endpoints of the conflicting edge (white vertices), which is given in sub-figure (c). In the \tc reduction matrix, only those entries will be assigned value $1$, for which there is an edge in sub-figure (c). The figure is best seen in color.}
\label{fig_sparse_graph_example}
\end{figure}

The distance matrix was computed using the \tc reduction, and consequently, many distances are `1'. Now, we first replace entries with value $1$ with the value $\vert V \vert$ unless the corresponding coefficient is that of an edge in $T$. Note that since, in general, $T$ is not a Hamiltonian cycle in $G$, there may be many conflicting edges (i.e., $e' \in E'-E$). We randomly select one conflicting edge $e'$ (if there are more than one conflicting edges) and replace it with the edges in the shortest path between the end-points of $e'$. We then iteratively remove all conflicting edges from the graph (add value $\vert V \vert$ in \tc reduction matrix for the endpoints $(x,y)$ of conflicting edges, i.e., $M_{x,y} = \vert V \vert$).
The step-by-step procedure to make the sparse graph is given in Figure~\ref{fig_sparse_graph_example}.

\subsubsection{Re-structuring the Population}
\begin{figure}[h!]
	\centering
	\includegraphics[page=8, scale=0.6] {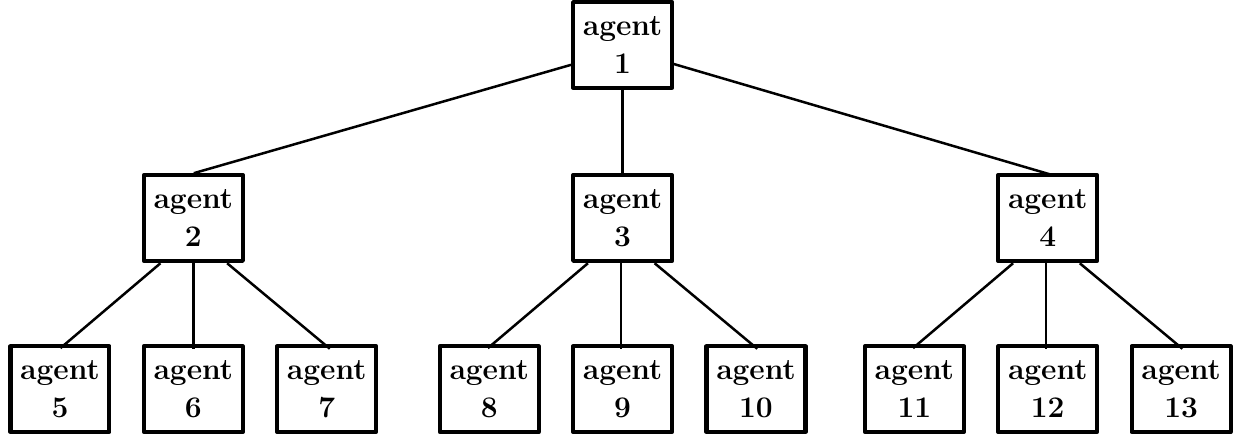}
	\caption{Agents population hierarchy.}
\label{fig_population_hirarchy}
\end{figure}

After the initialization of the population, we re-structure it using \textit{UpdatePocket} and \textit{PocketPropagation} procedures. These procedures are represented by \textit{structurePop()} procedure in Algorithm~\ref{algo_1}. We organize the population hierarchically as a complete ternary tree of $13$ agents (see Figure~\ref{fig_population_hirarchy}). These agents are clustered in $4$ sub-populations. These sub-populations are composed of four individuals ($1$ leader and $3$ children). Each agent maintains two solutions, namely \textit{pocket} and \textit{current}. The cost of the top pocket agent (agent $1$) is the minimum of the population. If the parent agent's pocket cost is greater than the pocket cost of the child, the solution parent and child are switched. This step is performed using \textit{PocketPropagation()} procedure.
Similarly, the cost of the pocket solution for an agent must be less than or equal to the cost of the same agent's current solution. If that is not the case, the solutions are again switched. This process is performed using \textit{UpdatePocket()} procedure.

\subsubsection{Recombination Procedure}
The generic pseudocode for the recombination procedure is given in Algorithm~\ref{algo_recombine}. In this algorithm, the recombination operator takes two feasible solutions (tours) $T_{1}$ and $T_{2}$ as input from population $P$ and returns another feasible solution $T_{off}$ (where $off$ refers to offspring).
The solution $T_{off}$ is generated using the Crossover procedure (discussed below), applying local search heuristic (RAI), and adding it to the population $P$.

We use the ``Strategic Arc Crossover (SAX)"~\cite{moscato1992memetic} operator to explore the search space of the instance by combining features from different agents in the population.
The selection of the feasible solutions ($T_1$ and $T_2$) for the SAX requires a suitable choice of agents to preserve population diversity. 
For each sub-population of the ternary tree given in Figure~\ref{fig_population_hirarchy} (a sub-population consists of a parent/leader and its three children/offspring), we define the following notation:
\begin{itemize}
    \item Leader: Parent node in a given sub-population
    \item $off_{1}$, $off_{2}$, and $off_{3}$: Three child agents of the Leader in the sub-population
    \item Pocket(i): Pocket solution of $i^{th}$ agent
    \item Current(i): Current solution of $i^{th}$ agent
\end{itemize}
The current of the root node is generated by giving the pocket solution of the first child and the current solution of the second child as input to \textit{Recombine Procedure} (see Algorithm~\ref{algo_recombine}). The recombination procedure work as follows for the rest of the sub-populations (for each node in the second level of the ternary tree in Figure~\ref{fig_population_hirarchy} and their children in level $3$):
\begin{itemize}
    \item Current (leader) $\leftarrow$ Recombine (Pocket ($off_{2}$), Pocket ($off_{3}$))
    \item Current ($off_{1}$) $\leftarrow$ Recombine (Pocket (leader), Current ($off_{2}$))
    \item Current ($off_{2}$) $\leftarrow$ Recombine (Pocket ($off_{1}$), Current ($off_{3}$))
    \item Current ($off_{3}$) $\leftarrow$ Recombine (Pocket ($off_{2}$), Current ($off_{1}$))
\end{itemize}
Note that the selection of children $off_{1}$, $off_{2}$, and $off_{3}$ is made randomly.

\begin{algorithm}[h!]
    \centering
    \caption{Recombination Procedure}
    \begin{algorithmic}[1]
        \State \textbf{Input:} Two tours $T_1$, $T_2$ $\in P$  \Comment{$P \rightarrow$ population}
        \State $G'$ := Crossover($T_1, T_2$) \Comment{SAX operator} 
        \State $T_{off}$ := Local-Search($G'$) 
        \State add $T_{off}$ to $P$
    \end{algorithmic}
    \label{algo_recombine}
\end{algorithm}

As mentioned before, to generate a new offspring, we use the SAX crossover operator within the recombination procedure. 
SAX crossover takes two tours $T_1$ and $T_2$ as input and generates a new graph $G'$. The $G'$ contains the edges from the union of $T_1$ and $T_2$ (without repetition). Each vertex in $G'$ will have $1$ or $2$ in-degree and out-degree along with multiple sub-tours (where a sub-tour is a path consisting of $\eta$ cities, the value of $\eta$ is $1 \leq \eta \leq N$, and $N$ is the total number of cities in the original graph). 
The pseudocode to generate the sub-tours using the SAX crossover is given in Algorithm~\ref{saxCrossover}.
At last, local search is applied on $G'$ to get a tour ($T_{off}$), and the solution is added to the population $P$. Algorithm~\ref{saxCrossover} works as follows:
\begin{enumerate}
    \item We pick a random city $c$ from the graph and mark it as visited. The city $c$ will be the initial city for the subtour. Now, $c$ will be the head and tail of the new subtour.
    \item While the current head has at least one out-neighbor that is unvisited, we do the following:
    \begin{enumerate}
        \item Randomly select one of those unvisited out-neighbor and add it to the end of the subtour.
        \item Declare it as new head
        \item Mark it as visited.
    \end{enumerate}
    \item While the current tail has at least one in-neighbor that is unvisited, we do the following:
    \begin{enumerate}
        \item Randomly select one of those unvisited in-neighbor and add it to the beginning of the subtour.
        \item Declare it as new Tail
        \item Mark it as visited.
    \end{enumerate}
    \item If there is any unvisited vertex remaining, we go to step 1.
\end{enumerate}

\begin{algorithm}[h!]
    \centering
    \caption{Create subTour using SAX crossover} \label{saxCrossover}
    \begin{algorithmic}[1]
        \State subTour = \{\} \Comment{linked list}
        \State citiesRemaining = $V(G)$ \Comment{set of all cities}
        \While{citiesRemaining $\neq$ empty}
            \State startCity = randomVertex($G$) \Comment{select unvisited city at random}
            \State removeCity(citiesRemaining,startCity) \Comment{marked startCity as used}
            \State updateHead(subTour,startCity) \Comment{declaring startCity as head of subTour}
            \State updateTail(subTour,startCity) \Comment{declaring startCity as tail of subTour}
            \State $vertesSet_{out}$ = $N_{out}$(subTour.head) $\in$ citiesRemaining \Comment{$N_{out} \rightarrow out-neighbors$}
                \While{$vertesSet_{out}$ $\neq$ empty}
                    \State $v$ = random($vertesSet_{out}$) \Comment{choose a random city}
                    \State insertAtEnd(subtour,$v$) \Comment{adding $v$ to the end of subtour}
                    \State removeCity(citiesRemaining,$v$) \Comment{marked $v$ as used}
                    \State updateHead(subTour,$v$) \Comment{declare $v$ as new head of subTour}
                \EndWhile
            \State $vertesSet_{in}$ = $N_{in}$(subTour.tail) $\in$ citiesRemaining \Comment{$N_{in} \rightarrow in-neighbors$}
                \While{$vertesSet_{in}$ $\neq$ empty}
                    \State $v$ = random($vertesSet_{in}$) \Comment{choose a random city}
                    \State insertAtStart(subtour,$v$) \Comment{adding $v$ to the start of subtour}
                    \State removeCity(citiesRemaining,$v$) \Comment{marked $v$ as used}
                    \State updateTail(subTour,$v$) \Comment{declare $v$ as new tail of subTour}
                \EndWhile
        \EndWhile
    \end{algorithmic}
\end{algorithm}

\subsubsection{Mutation Procedure}
\begin{figure}[h!]
	\centering
	\includegraphics[scale=0.3] {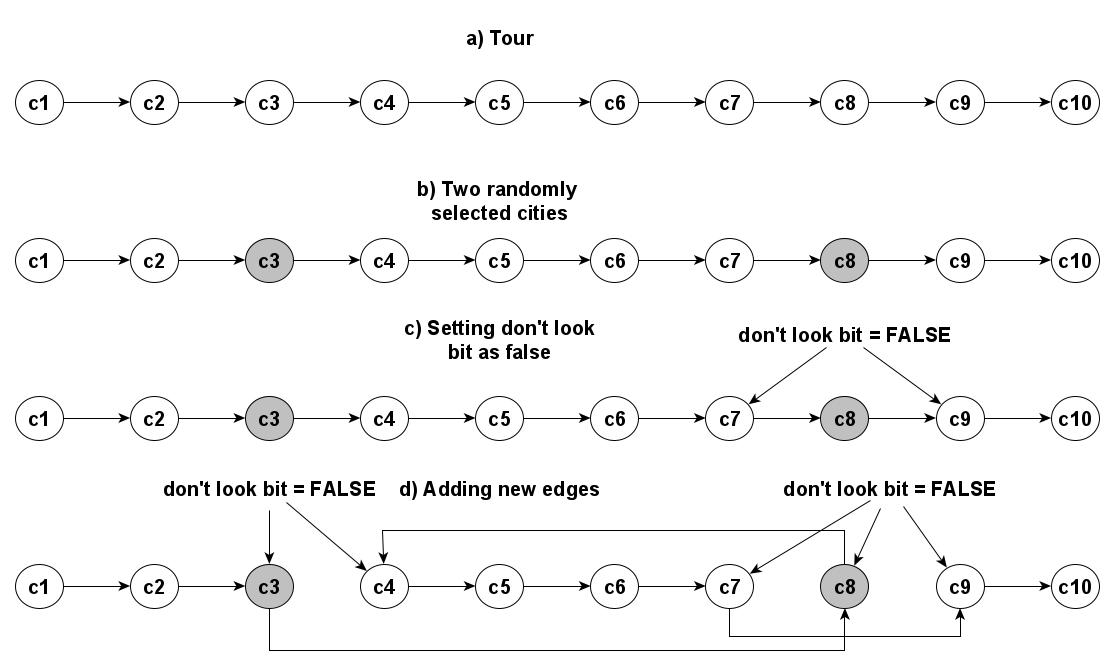}
	\caption{An example of how mutation works. Figure a) is a sample tour. In Figure b), we randomly select two cities, namely $c3$ and $c8$. In Figure c), we set don't look a bit as FALSE for previous and next neighbors of city $c8$. In Figure d), we include some extra edges (noise) in the tour to bring some diversity. We also set don't look a bit as FALSE for city $c8$, $c3$, and $c4$. We then remove the edges ($c3$,$c4$), ($c7$,$c8$), and ($c8$,$c9$).}
    \label{fig_mutation_example}
\end{figure}

The main idea behind the mutation procedure is to introduce some diversification in the population so that it may get out from a local optimum attractor region in configuration space~\cite{DBLP:journals/heuristics/BuriolFM04} (see Figure~\ref{fig_mutation_example}).  
The value for the mutation (mutation rate) is set to $0.05$ (selected empirically as in~\cite{DBLP:journals/heuristics/BuriolFM04}, it is also the same value is used in the literature~\cite{franca2006genetic}), which means that each new individual has a mutation probability of $5$\%. The mutation procedure is executed using \textit{mutatePop()} function. The pseudocode for the mutation procedure is given in Algorithm~\ref{algo_mutate}.

\begin{algorithm}[h!]
    \centering
    \caption{Mutation Procedure}
    \begin{algorithmic}[1]
    \For{$i \in P$} \Comment{$P \rightarrow Population$}
        \If{random() $\leq 0.05$} \Comment{$0.05 \rightarrow Mutation Rate$}
            \State aux = Mutate(i)
            \State aux = Local\_Search(aux) \Comment{\rai Local Search}
            \State $F_{P[i]}$ = cost(P[i]) \Comment{evaluate the total cost of current population solution}
            \State $F_{aux}$ = cost(aux) \Comment{evaluate the total cost of solution aux}
            \If{$F_{P[i]}$ $>$ $F_{aux}$}
            \State InsertIntoPopulation(aux, P) \Comment{replace P[i] with new solution ``aux"}
            \EndIf
        \EndIf
    \EndFor
    \end{algorithmic}
    \label{algo_mutate}
\end{algorithm}

In our \maa, we are using the concept of \textit{don't look bit} (introduced in~\cite{bentley1990experiments}) in recombination, mutation, and RAI local search.
This idea proved to be effective in order to reduce the search space for the local search algorithm (hence decreases the running time) without a loss in the quality of the solution.
Each city in the tour has its \textit{don't look bit} value set as TRUE or FALSE.
The cities that have their \textit{don't look bit} set as `FALSE' are considered as the \textit{critical} cities. These critical cities are used, for instance, as a starting point for the local search algorithm. 
In the mutation procedure, the five cities involved in the insertion move (in Figure~\ref{fig_mutation_example}) will have their don't look bit marked as FALSE. These $5$ cities will also have their don't look bit marked as FALSE for the execution of RAI local search. In the recombination procedure, all cities in the subtours (generated in SAX crossover as given in Algorithm~\ref{saxCrossover}) will have their don't look bit marked as FALSE.
After we apply the local search algorithm, we set \textit{don't look bit} as TRUE for all the cities.
For further detail regarding the concept of \textit{don't look bit}, we encourage the readers to follow~\cite{DBLP:journals/heuristics/BuriolFM04}. 

\subsubsection{Restart Population}
There may come a scenario where the population loses its diversity. In this case, we aim to restart the search (if the number of generations exceeds $30$) by re-initializing the population (to create the new population) using the nearest neighbor heuristic. We take the best solutions found so far (in the population) and add them unaltered in the new population. We then mutate and optimize the current and pocket solutions (except the best solutions) using \rai local search heuristic. The pseudocode for \textit{restartPop()} is given in Algorithm~\ref{algo_restart_pop}.
\begin{algorithm}[h!]
    \centering
    \caption{Restart Population Procedure}
    \begin{algorithmic}[1]
        \If{P converged} \Comment{$P \rightarrow$ population}
        \State P = reinitialise(P) \Comment{using nearest neighbor heuristic}
            \For{$i \in P.current$}
               \State  $i :=$ Local-Search(Mutate(i)) \Comment{mutate and optimize the currents}
            \EndFor
            \For{$i \in P.pocket \setminus {best}$}
               \State  $i :=$ Local-Search(Mutate(i)) \Comment{mutate and optimize pockets except  best pocket}
            \EndFor
        \EndIf
    \end{algorithmic}
    \label{algo_restart_pop}
\end{algorithm}

\subsubsection{Local Search}\label{subsection_local_search}
\begin{figure}[h!]
	\centering
	\includegraphics[scale=0.24] {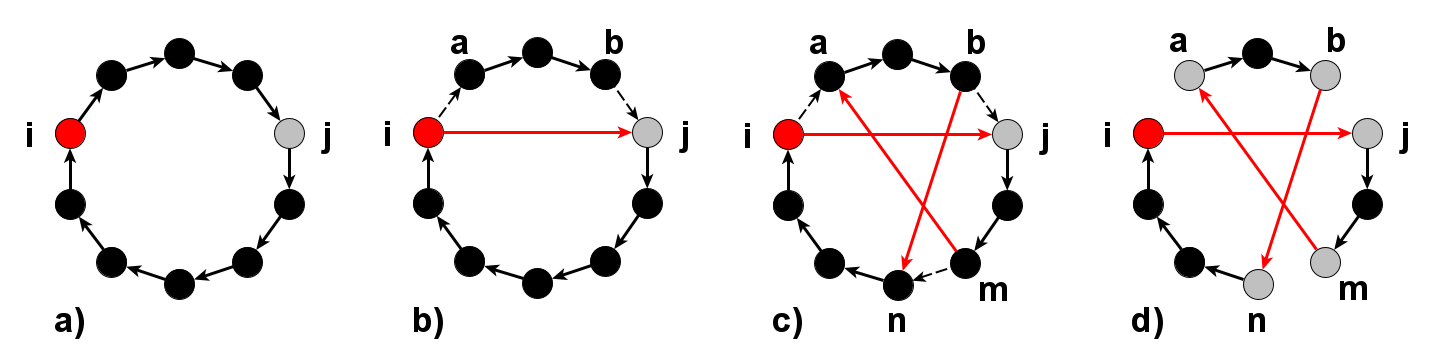}
	\caption{\rai local search movements. a) We select a critical city $i$ and a city $j$, where $j$ is p-in-neighbour of $i$ (see section~\ref{subsection_local_search} for definition of p-in-neighbour). b) An edge is inserted between $i$ and $j$ and we also identify two cities $a$ and $b$. c) Two cities $m$ and $n$ are selected, and edges are inserted between $(m,a)$ and $(b, n)$. d) Remove edges $(i, a)$, $(b, j)$, and $(m, n)$.}
\label{fig_rai_local_search}
\end{figure}

After \textit{mutatePop()} procedure, we apply a previously proposed local search method for the ATSP called \rai for each agent separately (using \textit{optimizePop()} function), which requires an ordered list of $5$ nearest neighbors of each city. The main idea of \rai is to improve the tours starting from cities marked with \textit{don't look bits}~\cite{bentley1990experiments} during the recombination and mutation steps. In this local search, the basic movement is of the $3$-Opt type. 
A movement of the \rai local search is illustrated in Figure~\ref{fig_rai_local_search}. 
Consider a critical city $i$ (having don't look bit marked as FALSE). Two types of neighbors are defined in \rai, namely $s$-out-neighbours(i) and $p$-in-neighbors(i). The first type consists of a set of neighbors (destination cities) of the $s$ shortest outgoing edges from the city $i$. The second type consists of a set of neighbors that are constructed by the set of the cities of the $p$ shortest incoming edges to the city $i$. For a given tour, suppose $i$ is a critical starting city. The \rai then works as follows:
\begin{enumerate}
    \item Suppose $j$ is the $p$-in-neighbors(i) (see Figure~\ref{fig_rai_local_search} (a)). 
    \item Now insert an edge between $i$ and $j$ (see Figure~\ref{fig_rai_local_search} (b)) and identify the edges ($i,a$) and ($b,j$), where $a = Next(i)$ and $b = Prev(j)$. Calculate $\Delta_1 = d_{ia} + d_{bj} - d_{ij}$.
    \item Select an edge (m,n) starting from (j, Next(j)) (where $(m,n) \neq (i,j)$) such that $\Delta_2 < \Delta_1$ (where $\Delta_2 = d_{ma} + d_{bn} - d_{mn}$). If such an edge $(m,n)$ is found, included two new edges ($m,a$) and ($b,n)$ (see Figure~\ref{fig_rai_local_search} (c)).
    \item Now remove the edge ($m,n$), ($i,a$), and ($b,j$) (see Figure~\ref{fig_rai_local_search} (d)).
\end{enumerate}
For a detailed explanation regarding RAI's working, we refer the readers to the work done in~\cite{DBLP:journals/heuristics/BuriolFM04}.

After applying \rai to each agent, a more sophisticated local search method (\lkh heuristic) is separately used in order to optimize the results of each agent. 
We found through initial experimentation that the combination of \rai and \lkh yields better results than using any of these methods alone, so for simplicity, we only present these results.

\subsubsection{Augmentation via the shortest paths to avoid conflicting edges}
Since we started with the sparse version of the graph, it may not be Hamiltonian. 
In Figure~\ref{fig_sparse_graph_example} (a), we have an original graph. To make this graph sparse, we apply a \lkh local search step. It gives us a tour with $\geq 1$ conflicting edges. We use that tour (with conflicting edges) as a sparse graph (as given in Figure~\ref{fig_sparse_graph_example} (b)). To deal with the conflicting edge problem, we replace these conflicting edges with the set of edges in the shortest path between the endpoints of the conflicting edges, as given in Figure~\ref{fig_sparse_graph_example} (c).
Since the graph may still not be Hamiltonian, this means that we need to include more entries with value $1$ in the \tc matrix. In this way, if the solution is not already found with the current matrix, a new set of coefficients needs to be set to `1', which may help the \ma to find a solution.
To do this, we run the \lkh heuristic on the sparse version of \tc matrix (on the top/leader agent), which will return us a tour $T''$ (this is done using \textit{sparsification()} procedure in Algorithm~\ref{algo_1}). Note that there may be one or more conflicting edges in $T''$. 
For any edge of cost greater than 1 in $T''$, i.e., corresponding to a conflicting edge $e'$ from $T''$ (where $e' \notin E$, see Definition~\ref{def_conflicting}), we identify the set of edges in the shortest path (in the original graph $G$) between the end-points of $e'$.
For all the corresponding entries of edges of this path in the \tc matrix, we will now assign the value $1$ to their coefficients. In this way, in each iteration of the generation loop, some new edges (new entries with value $1$) will be added in the sparse \tc matrix.

\subsubsection{Reset}
The  \textit{sparsification()} procedure above keeps changing entries to the value of `1' (in \tc matrix), and there will come to a point where the number of edges (number of $1's$) in the sparse \tc matrix becomes equal to the number of $1's$ in the original (non-sparse) \tc matrix. At that point, we reset the entries with value $1$ in the sparse \tc matrix using \textit{initialSparsification()} procedure. 
All the other (remaining) entries with value $1$ (which were not marked as $1$ by \textit{initialSparsification()} procedure) are replaced with the value equal to the distance between the respective endpoints (i.e., a classical \tc reduction approach).
Hence, at no stage do we allow the sparse matrix to operate with the number of $1$'s equal to the total number of $1$'s in the original (non-sparse) \tc matrix. 
In this heuristic way, we expect to be working with graphs having a lower treewidth. 

\subsubsection{Motivation for the augmentation heuristic}

It is well known that if a greater number of edges exist in the graph, then there is a high chance of the presence of \hc in the graph due to two Hamiltonian sufficiency theorems: 
\begin{theorem}\label{theorm_hc}
Let $G = (V,E)$ be the simple undirected graph, where $\vert V \vert > 3$. For every pair of distinct non-adjacent vertices $a$ and $b \in V$, if $degree(a) + degree(b) \geq \vert V \vert$, then we can say that $G$ is Hamiltonian. This is known as Ore’s theorem~\cite{ore1960note}.
\end{theorem}
\begin{theorem}\label{theorm_hc2}
Let $G = (V,E)$ be the simple undirected graph, where $\vert V \vert > 3$. If every vertex $v \in V$ has degree $\geq \frac{\vert V \vert}{2}$, then we can say that $G$ is Hamiltonian. This is known as Dirac's Theorem~\cite{dirac1952some}.
\end{theorem}

However, these are high bounds. An increased number of edges means that the treewidth of the graph will (most probably) be higher, which means that the underlying algorithm or heuristic has to put extra effort (hence increased runtime) to find the \hc of length $\vert V \vert$~\cite{ziobro2019finding}. 
This said, in order to simplify the task of the method, not only the sparsification of the graph but also the augmentation becomes necessary.

\subsection{Stopping criteria} \label{sec_stopping_criteria}
There are three conditions to stop the \maa.
\begin{enumerate}
    \item The first one is related to the length of the tour (local optimal solution). If any agent of the memetic algorithm finds a \hc of length $n$ (where $n = \vert V \vert$), the algorithm terminates since we have found the \hc of $G$. 
    \item The second criterion is related to the runtime. We stop the \ma when it is unable to find the \hc in a predefined time ($30$ minutes for the Set C and Set F instances, and $10$ minutes for set A instances ``see Section~\ref{dataset_statistics} for more detail"). We note that increasing the assigned runtime did not contribute a great deal towards solving more instances.
    \item The third criterion is related to the number of generations of the memetic algorithm. If the number of generations in the memetic algorithm becomes equal to the maximum allowed generations, the algorithm terminates. The maximum number of generations is given in Equation~\ref{eq_max_generations}, which was introduced in~\cite{DBLP:journals/heuristics/BuriolFM04}.
    \begin{equation}\label{eq_max_generations}
        \text{Max Generations = } 5 \times 13 \times log(13) \times \sqrt{n}
    \end{equation}
    where $5$ is the mutation rate, $13$ represents the number of agents in the ternary tree, and $n$ is the number of cities in a given instance.
\end{enumerate}

\subsection{Parameter Values}
The values for all parameters used in our \ma are given in Table~\ref{tbl_parameters}.
In~\cite{DBLP:journals/heuristics/BuriolFM04} the
authors show results with other parameters (e.g., population size, topology of the tree, mutation rate, etc.). Whenever possible, we set the same parameters as well. We refer to Section~\ref{dataset_statistics} for the classification of the instances in three groups that motivated different running times. 

\begin{table}[h!]
\centering
\caption{Parameter values of the \ma to solve the \tsp instances.}
\label{tbl_parameters}
\resizebox{0.99\textwidth}{!}{
\begin{tabular}{lc}
\toprule
Parameters & Values \\
\midrule
Mutation Rate & 0.05 \\
Max Generations &  $5 \times 13 \times log(13) \times \sqrt{n_{cities}}$\\ 
Population Size (ternary tree) & 13 \\
Max Runtime for Set A instances & 10 minutes \\
Max Runtime for Set C, Set F instances & 30 minutes \\
Number of generations to reset population & $30$ \\
Number of conflicting edges replaced with a shortest path & $1$ \\
\bottomrule
\end{tabular}
}
\end{table}

\section{Experimental Setup}\label{sec_experimental_setup}
In this section, we present dataset description, detail of baseline methods, evaluation metric, and implementation detail of our proposed \maa.

\subsection{Dataset Description}\label{dataset_statistics}
We are using the well-known set of \hc instances from {\em Flinders University Hamiltonian cycle Problem Challenge Set} (\fhcpsc)~\cite{Haythorpe2015}, which consist of $1001$ instances, all of them are Hamiltonian graphs. We follow the classification of~\cite{ziobro2019finding} that used the notion of treewidth to separate the \hcp instances in different sets. They have partitioned the \fhcpscc instances into \textit{`Set A'} (all the graphs with small treewidth, $623$ instances, treewidth at most $8$), and \textit{`Set C'} (a set of larger treewidth graphs, $19$ instances, having treewidth between $19$ and $42$). We note that they used several instances for tuning their algorithm (called \textit{`Set B'}, which we have included in the \textit{Set A} since we do not need to tune parameters in our approach. 
We include all the remaining instances (which are not included in set A and set C) into a new group \textit{`Set F'} that has $359$ graphs. The dataset descriptive statistics are given in Table~\ref{tbl_dataset_statistics}. Again, we include a set of instances called \textit{`Set B'} in~\cite{ziobro2019finding} as well in \textit{`Set A'}.

\begin{figure}[h!]
\minipage{0.48\textwidth}
  \includegraphics[height = 6cm,width=\linewidth] {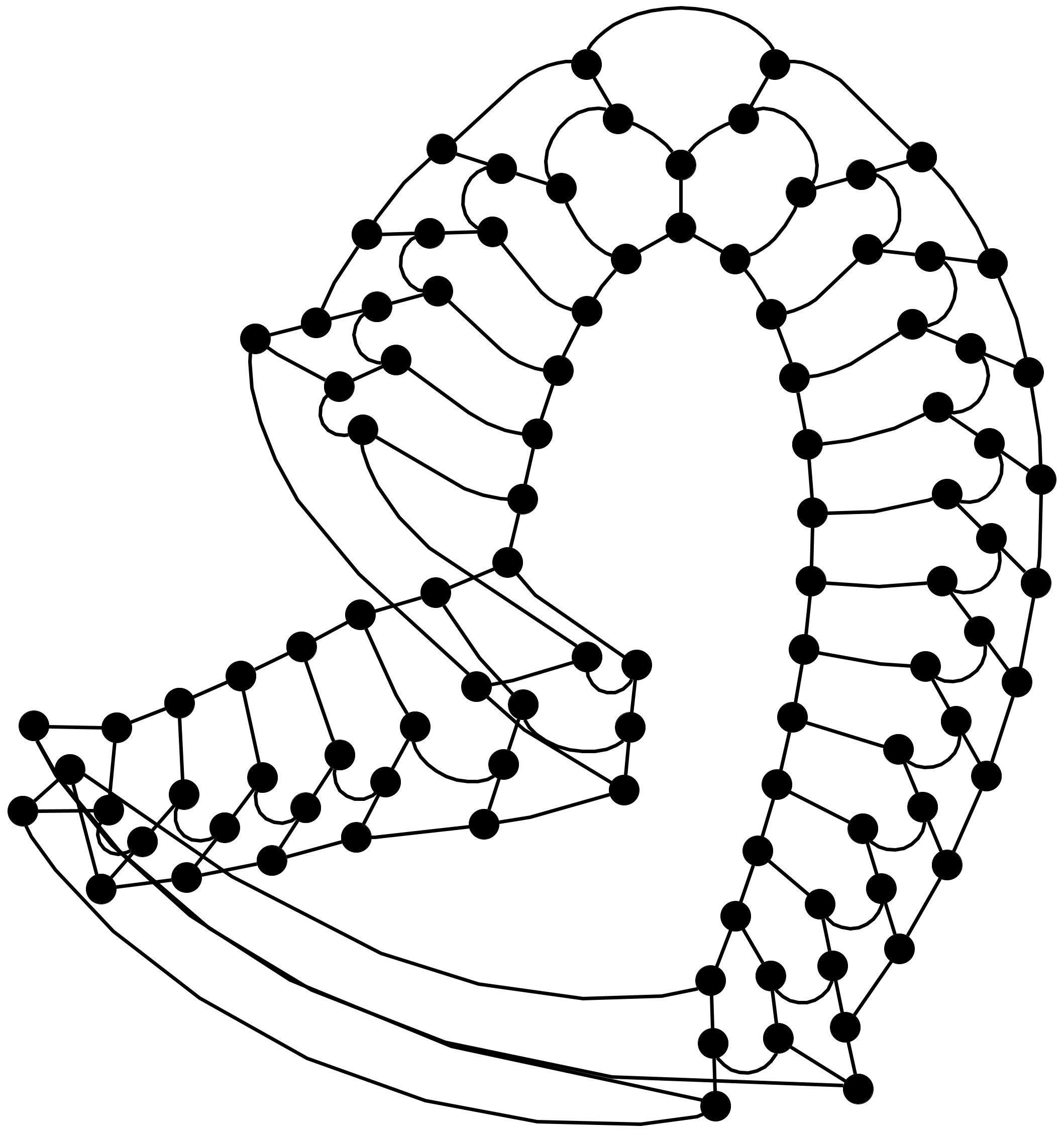}
  \caption*{(a) Instance 7}
\endminipage\hfill
\minipage{0.48\textwidth}
  \includegraphics[height = 6cm,width=\linewidth] {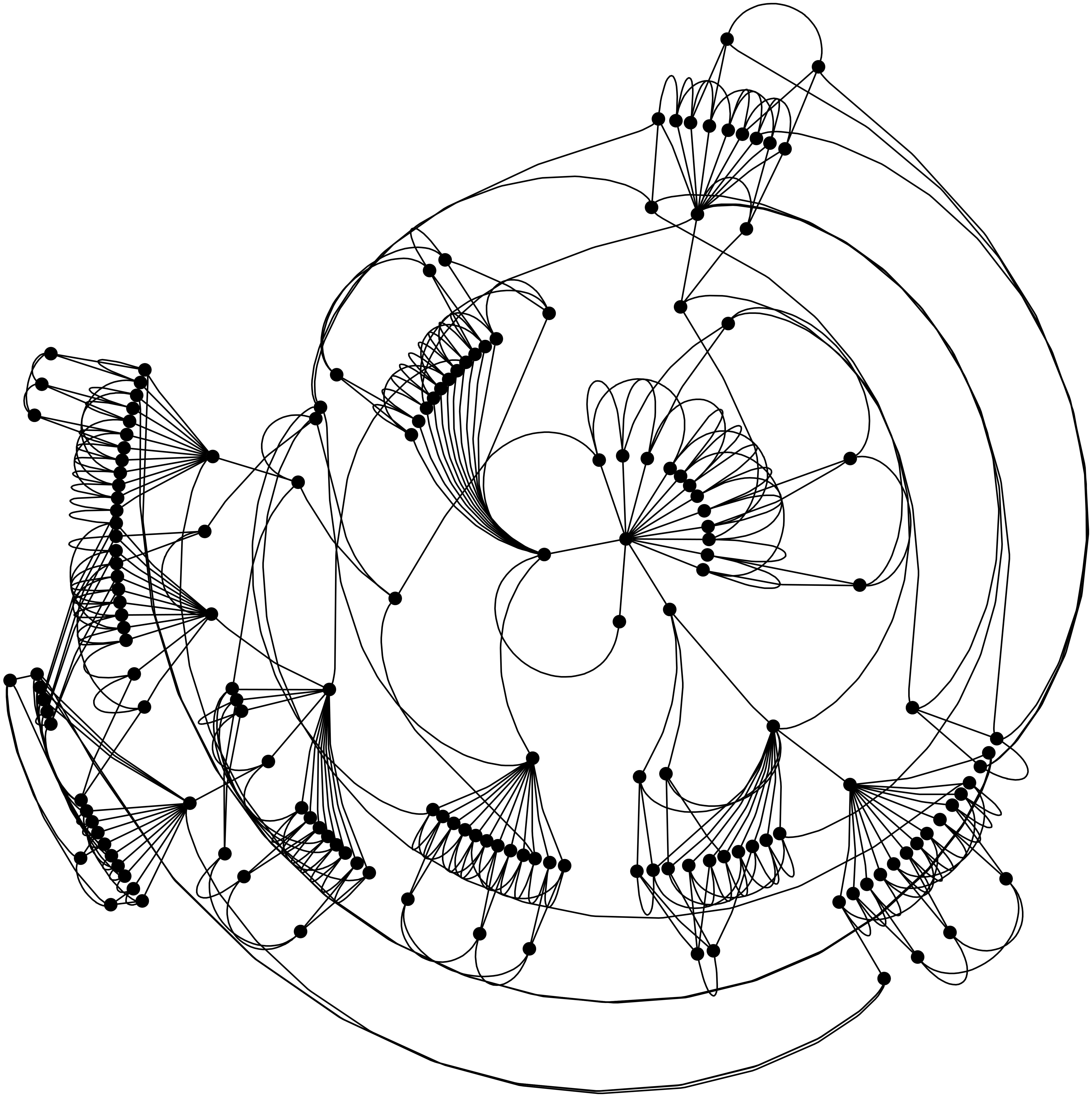}
  \caption*{(b) Instance 19}
\endminipage\hfill
\caption{Illustrative examples of different graph types in \fhcpscc (drawings produced using the radial layout option of the software \texttt{yEd}). (a) is an example of $3$-regular graph (instance $7$ with treewidth $9$). (b) is an example of a more complex structure, a graph that is not $3$-regular
(instance $19$ with treewidth $7$).
It is evident from the structures of these two graphs that (a) is easier to solve than (b) for a treewidth-based strategy, but for our method, both are solved in less than a second. In contrast with our results, we note that the Concorde-only based approach of~\cite{DBLP:conf/ssci/MathiesonM20} fails to solve instance $19$ under $10$ minutes.}
\label{fig_graph_struct_regular_vs_non}
\end{figure}

We note that alternative classifications are also possible as several instances of the \fhcpscc share a common structure. For instance, among the first $20$ instances in lexicographic order in the database, all of them are $3$-regular graphs except instance $19$. The structure of instance $7$ and instance $19$ are shown in Figure~\ref{fig_graph_struct_regular_vs_non}. However, we choose to follow the classification of instances as proposed in~\cite{ziobro2019finding} in order to make a fair and more straightforward comparison with their results.

\begin{table}[h!]
\centering
\caption{Dataset Statistics.}
\label{tbl_dataset_statistics}
\resizebox{0.99\textwidth}{!}{
\begin{tabular}{@{\extracolsep{4pt}}ccccccccc@{}}
\toprule
\multirow{2}{*}{Set} & \multicolumn{4}{c}{$\vert V(G) \vert$}  & \multicolumn{4}{c}{$\vert E(G) \vert $} 
\\
\cmidrule{2-5} \cmidrule{6-9}
 & Min & Avg & Median & Max & Min & Avg & Median & Max \\
\cmidrule{1-1} \cmidrule{2-5} \cmidrule{6-9}
A & 66 & 2,406.98 & 2,224 & 8,886 & 99 & 5,246.36 & 3,871 &  35,018 \\
C & 462 & 2,173.05 & 1,578 & 9,528 & 756 & 3,380.16 & 2,688 &  13,968 \\
F & 400 & 4,334.98 & 4,430 & 8613 & 825 & 11,131.44 & 7,643 & 31,5283 \\
\midrule

\midrule
\multirow{2}{*}{Set}  & \multicolumn{4}{c}{Minimum Degree}  & \multicolumn{4}{c}{Average Degree}
\\
 \cmidrule{2-5} \cmidrule{6-9}
  & Min & Avg & Median & Max & Min & Avg & Median & Max \\
 \cmidrule{1-1} \cmidrule{2-5} \cmidrule{6-9}
A & 2 & 2.83 & 3 & 4 & 3 & 3.95 & 3 & 8.73\\
C & 2 & 2 & 2 & 2 & 2.93 & 3.20 & 3.17 & 3.47\\
F & 2 & 2 & 2 & 2 & 2.79 & 13.41 & 3.39 & 561.50 \\
\midrule

\midrule
\multirow{2}{*}{Set}  & \multicolumn{4}{c}{Maximum Degree}  & \multicolumn{4}{c}{Girth} 
\\
 \cmidrule{2-5} \cmidrule{6-9}
  & Min & Avg & Median & Max & Min & Avg & Median & Max \\
  \cmidrule{1-1} \cmidrule{2-5} \cmidrule{6-9}
A & 3 & 159.61 & 4 &  1,908 & 3 & 3.29 & 3 & 5\\
C & 3 & 69.16 & 8 &  192 & 3 & 3.95 & 4 & 4 \\
F & 3 & 82.71 & 17 & 1122 & 3 & 3.40 & 3 & 6 \\
\midrule

\midrule
\multirow{2}{*}{Set}  & \multicolumn{4}{c}{Diameter}  
& \multicolumn{4}{c}{Density}
\\
 \cmidrule{2-5} 
 \cmidrule{6-9}
  & Min & Avg & Median & Max & Min & Avg & Median & Max \\
 \cmidrule{1-1} \cmidrule{2-5} 
 \cmidrule{6-9}
A & 6 & 194.13 & 89 & 1,113 & 0.0003 & 0.0044 & 0.0018 & 0.5012 \\
C & 10 & 24.26 & 18 & 92 & 0.0003 & 0.0027 & 0.0021 & 0.0070 \\
F & 2 & 9.16 & 9 & 14 & 0.0003 & 0.0148 & 0.0007 & 0.5010 \\

\bottomrule
\end{tabular}
}
\end{table}

\subsection{Baseline Methods used for comparison}
We are using the following baseline methods for comparison purposes since the ones that performed the best in this challenging dataset:
\begin{itemize}
    \item \textbf{Na\"{\i}ve Algorithm:}~\cite{bodlaender2015deterministic} Given a tree decomposition of the input graph of width $t$, this method uses dynamic programming algorithm to solve the \hcp in $2^{O(t \ log \ t)}$ time. See section~\ref{subsection_naive_algorithm} for detail.
    \item \textbf{HybridHAM:}~\cite{seeja2018hybridham}
    This algorithm proposes an efficient hybrid heuristic, which combines greedy, rotational transformation, and unreachable vertex heuristics to solve the \hcpp. Since they have not made their source code publicly available, we use the results reported in their paper for comparison with our proposed approach.
    \item \textbf{Rank $1$:}~\cite{ziobro2019finding} This method (also called {\em rank-based approach}), is a na\"{\i}ve approach with pruning of the state space leading to $4^{t}$ size bound. 
    \item \textbf{Rank $2$:}~\cite{ziobro2019finding,cygan2018fast} This technique is a modified version of the rank-based approach with the improved basis yielding the size bound $(2 + \sqrt{2})^{2}$. 
    \item \textbf{Concorde:}~\cite{DBLP:conf/ssci/MathiesonM20} This technique first uses different reduction techniques to reduce the Hamiltonian cycle problem instance to one of the traveling salesman problems. Then Concorde \tsp solver~\cite{Concorde} is applied on these reduction matrices to compute the \hcc. Among the proposed reduction methods, transitive closure (TC) appeared to be the most effective. Therefore, we use the same \tc reduction method and apply the Concorde solver to compute the results for this baseline method. Since they have reported results for the first $100$ instances only, we will compare results only for these $100$ instances.
\end{itemize}

The winner of the \fhcpscc solved $985$ instances (see Section~\ref{sec_most_solved_instances} for detail). However, to the best of our knowledge, no theoretical and experimental detail of the winner's procedure is available online (no results for individual instances are available, and no algorithm has been specified). Therefore, it is not possible to compare our \ma with the results of the winner of the challenge because there is no single algorithmic framework for the approach~\ref{sec_most_solved_instances}.

The codes for Na\"{\i}ve Algorithm, Rank $1$, and Rank $2$ method are available online from~\footnote{\url{https://github.com/stalowyjez/hc_tw_experiments}} and the code for Concorde is available online from~\footnote{\url{http://www.math.uwaterloo.ca/tsp/concorde.html}}, which has greatly help ed us to compare with our method and to ensure reproducibility across different computing platforms. We are grateful to all these researchers for making them available. 

\subsection{Evaluation Metric}
We use the runtime (in seconds) to find the \hc of length $\vert V \vert$ as an evaluation metric to measure the performance of the proposed \maa. The timeout is set to $10$ minutes for set $A$ instances while $30$ minutes for all other instances because the same timeout values have been used in the literature before~\cite{ziobro2019finding,DBLP:conf/ssci/MathiesonM20}.

\subsection{Computer programs developed and other software and hardware used}

All experiments were carried out on a Windows $10$ machine with an Intel(R) Core i3 CPU processor at $2.6$ GHz and $4$GB of DDR$3$ memory. The code for reduction-based methods is implemented in \textit{R}, and the code for the memetic algorithm is implemented in Java.
The code for \lkh is from Helsgaun's famous implementation and enhancements~\cite{HELSGAUN2000106}. To compute the treewidth of the graphs, we used the heuristic introduced in~\cite{hamann2016graph}, which is available online~\footnote{\url{https://github.com/kit-algo/flow-cutter-pace17}}.

\section{Results and Discussion} \label{results_and_discussion}
In this section, we report and discuss the time taken by our algorithm to find the \hc in the first $100$, set A, set C, and set F instances separately, and compare the results with those computed for different baseline methods.

\subsection{Comparison for solving most instances} \label{sec_most_solved_instances}
The overall results statistics for all instances of \fhcpscc are shown in Table~\ref{tbl_overall_results_stats}. We can observe that in terms of solving most instances, the proposed model outperforms all the baseline methods on the first $100$, set $A$, set $C$, and set $F$ instances.

According to~\cite{haythorpe2019fhcp}, the people securing top $5$ positions in \fhcpscc competition solved the following number of instances:
\begin{itemize}
    \item N. Cohen and D. Coudert (INRIA, France), $985$ graphs solved
    \item A. Johnson (IBM, United Kingdom), $614$ graphs solved
    \item A. Gharbi and U. Syarif (King Saud University, Saudi Arabia), $488$ graphs
    \item M. Noisternig (TU Darmstadt, Germany), $464$ graphs solved
    \item M. Nurhafiz (Independent Researcher), $385$ graphs solved
\end{itemize}
By analyzing the number of instances solved by the top $5$ candidates/groups, we observe that we solved more instances than the person in second place (see Table~\ref{tbl_overall_results_stats}). The `$\_$' sign in Table~\ref{tbl_overall_results_stats} means that the given approach is unable to solve any instance of the set.
\begin{table}[h!]
\centering
\caption{Overall result statistics for finding \hc (solved instances) and comparison with state-of-the-art solvers (which present results for all instances). The `$\_$' sign indicates that the given algorithm is unable to solve any instance of the set.}
\label{tbl_overall_results_stats}
\resizebox{0.99\textwidth}{!}{
\begin{tabular}{lcccc}
\toprule
& This approach & Na\"{\i}ve Algorithm & Rank 1 &  Rank 2 \\
\midrule
First 100 & \textbf{100} & 92 & 91 & 91 \\
Set A (623 instances)  & \textbf{596}  & 574  & 575  & 581 \\
Set C (19 instances)  & \textbf{16}  &  5 & 4  & 6 \\
Set F (359 instances)  & \textbf{43}  & \_  & \_  & \_ \\
\midrule
Total Solved (1001 instances) &  \textbf{655} & 579  &  579 & 587 \\
\bottomrule
\end{tabular}
}
\end{table}
\subsection{Comparison based on Runtime}
\begin{figure}[h!]
	\centering
	\includegraphics[page=7, scale=0.55] {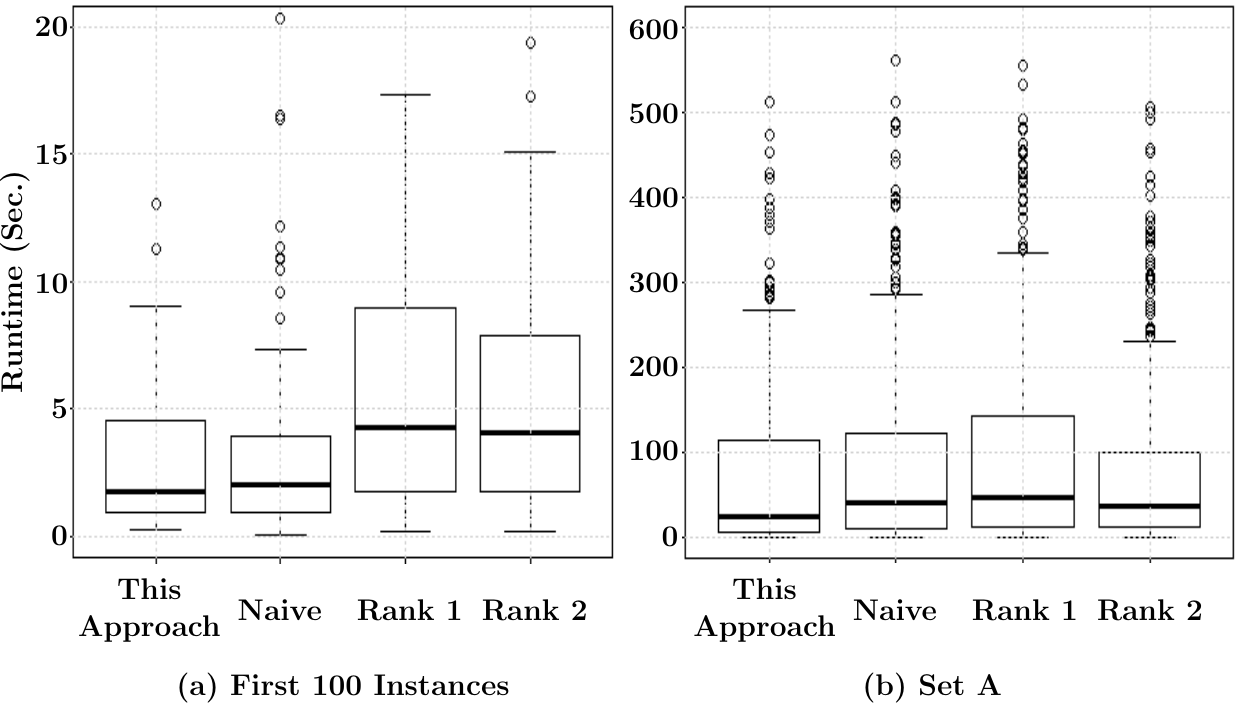}
	\caption{Comparison of runtime (for instances solved by all methods) of different methods on first 100 and set A instances.}
\label{fig_set_A_first_100}
\end{figure}

The boxplot comparison of runtime for different approaches on the first $100$ and set A instances is shown in Figure~\ref{fig_set_A_first_100}. We can observe that the proposed method is comparatively better than the Rank 1 and Rank 2 approach for the first $100$ instances while slightly worse than the Na\"{\i}ve Algorithm in terms of runtime. However, for the set A instances, the proposed method is better than the Na\"{\i}ve Algorithm and Rank $1$ approach. However, it is slightly worse than the Rank $2$ algorithm. Hence, all algorithms seem relatively in terms of runtime performance.
However, out of $623$ instances in set A, our method solved $596$ instances, Na\"{\i}ve Algorithm is able to find \hc in $574$ instances, Rank 1 solved $575$ instances, while Rank 2 solved $581$ instances. 
Therefore, we can conclude that our algorithm is comparable with baselines in terms of runtime; however, in terms of finding \hcc, the proposed approach outperforms the baselines for the instances of set A.

Table~\ref{tbl_treewidth_set_c} shows the runtime comparison of different methods for set C instances. We can observe that our proposed method outperforms the existing methods for most instances of set C. Our algorithm is better not only in terms of runtime but also in finding \hc for most instances. Most baseline models solved the instances with comparatively smaller treewidth (i.e., $\leq 28$). However, our proposed \ma solved instances with greater treewidth.

\begin{table}[h!]
\centering
\caption{Runtime comparison with methods proposed in~\cite{ziobro2019finding} and~\cite{DBLP:conf/ssci/MathiesonM20} on \textit{Set C} instances. Time is measured in seconds. The term ``Total" means the number of solved instances. In the table, TW represents the treewidth.}
\label{tbl_treewidth_set_c}
\resizebox{0.99\textwidth}{!}{
\begin{tabular}{ccccccccc}
\toprule
Inst. \# & {$\vert V \vert$} & {$\vert E \vert$} & {TW} & {This approach} & Na\"{\i}ve Algo. & Rank 1 & Rank 2 & Concorde \\
\midrule
74 & 462 & 756 & 34 & 391.02 & \_ & \_ & \_ & \textbf{4.52}\\
109 & 606 & 933 & 20 &  25.44 & \textbf{1.00} & 2.20 & 0.90  & 2.47 \\
110 & 606 & 925 & 19 &  30.06 & \textbf{0.67} & 1.03 & 0.70  & 14.98 \\
144 & 804 & 1256 & 23 & 27.44 & 57.35 & 486.81 & 58.78  & \textbf{14.45} \\
145 & 804 & 1252 & 23 &  \textbf{1.22} & 476.07 & 504.85 & 150.9  & 30.97 \\
172 & 1002 & 1575 & 28 &  \textbf{6.22} & 401.23 & \_ & 38.20  & 11.09 \\
173 & 1002 & 1579 & 26 &  \textbf{15.27} & \_ & \_ & 128.9 & 29.17 \\
199 & 1200 & 1902 & 32 &  14.13 & \_ & \_ & \_  & \textbf{6.72} \\
200 & 1200 & 1902 & 30 &  \textbf{22.05} & \_ & \_ & \_  & 469.12 \\
253 & 1578 & 2688 & 41 &  \textbf{0.64} & \_ & \_ & \_  & 482.45 \\
268 & 1644 & 2767 & 42 &  \textbf{0.85} & \_ & \_ & \_  & 492.17 \\
271 & 1662 & 2770 & 39 &  \textbf{10.56} & \_ & \_ & \_  & \_ \\
272 & 1662 & 2863 & 36 &  \textbf{23.21} & \_ & \_ & \_  & \_ \\
290 & 1770 & 3020 & 37 &  \textbf{18.09} & \_ & \_ & \_  & \_ \\
298 & 1806 & 3071 & 33 &  \textbf{37.93} & \_ & \_ & \_  & \_ \\
340 & 2010 & 3488 & 31 &  \textbf{72.31} & \_ & \_ & \_  & \_ \\
703 & 4024 & 5900 & 25 &  \_ & \_ & \_ & \_  & \_ \\
989 & 7918 & 11608 & 25 &  \_ & \_ & \_ & \_  & \_ \\
1001 & 9528 & 13968 & 26 &  \_ & \_ & \_ & \_  & \_ \\
\midrule
Total & \_ & \_ & \_ & \textbf{16} & 5 & 4 & 6  & 11 \\
\bottomrule
\end{tabular}
}
\end{table}

\subsubsection{Generalization of our \ma on larger and difficult instances}
In this section, we investigate the performance of our proposed \ma on comparatively more difficult instances, i.e., those that have higher treewidth and present more challenging structures. These sets of instances are categorized into a separate set called `Set F' (see Section~\ref{dataset_statistics}).
The runtime for the solved instances of Set F is shown in Table~\ref{tbl_treewidth_set_f}. For the Set F instances, Na\"{\i}ve Algorithm, Rank $1$, and Rank $2$ methods were unable to solve even one instance. Hence, the proposed model is the dominant one for these instances. We can observe in Table~\ref{tbl_treewidth_set_f} that some instances have treewidth greater than $200$, but still, our \ma was able to find the \hc in those instances. This behavior shows that our algorithm can also find Hamiltonian cycles in graphs with greater treewidth.

\begin{table}[h!]
\centering
\caption{Runtime (in sec.) for Set F instances (which are solved by our \maa). The term ``Total" means the number of solved instances. In the table, TW represents the treewidth.}
\label{tbl_treewidth_set_f}
\resizebox{0.99\textwidth}{!}{
\begin{tabular}{@{\extracolsep{4pt}}
p{0.5cm}cccp{1.8cm}
p{0.5cm}cccp{1.8cm}
@{}}
\toprule
Inst. \# & {\multirow{2}{*}{$\vert V \vert$}} & {\multirow{2}{*}{$\vert E \vert$}} & {\multirow{2}{*}{TW}} & {This Approach} & Inst. \# & {\multirow{2}{*}{$\vert V \vert$}} & {\multirow{2}{*}{$\vert E \vert$}} & {\multirow{2}{*}{TW}} & {This Approach}\\
\cmidrule{1-5}\cmidrule{6-10}
72  & 460 & 52901 & 231 & 41.55 & 
251 & 1566 & 2773 & 46  & 1670.50 \\
73 & 462 & 693 & 9 & 3.30 & 
255 & 1584 & 2799 & 45  & 739.28 \\
79 & 480 & 57601 & 241 & 123.79 & 
257 & 1596 & 2564 & 38  & 1510.39 \\
84 & 500 & 62501 & 251 & 84.54 & 
258 & 1596 & 2556 & 42  & 1276.50 \\
90 & 510 & 65026 & 256 & 61.70 & 
263 & 1620 & 2820 & 40 & 545.66 \\
96 & 540 & 72901 & 271 & 281.15 & 
264 & 1626 & 2820 & 45 & 834.97 \\
98 & 546 & 825 & 36 & 83.32 & 
284 & 1746 & 3009 & 43 & 1617.40 \\
128 & 677 & 114583 & 339 & 78.63 & 
286 & 1758 & 3127 & 44 & 1037.64 \\
130 & 690 & 979 & 37 & 75.90 & 
294 & 1794 & 2899 & 42 & 650.13 \\
134 & 724 & 131045 & 363 & 198.90 & 
295 & 1794 & 2911 & 44 & 310.05 \\
149 & 819 & 1278 & 56 & 51.03 & 
297 & 1806 & 3119 & 42 & 1732.97 \\
150 & 823 & 169333 & 412 & 453.81 & 
315 & 1896 & 3259 & 47 & 1191.38 \\
162 & 909 & 206571 & 455 & 66.75 & 
323 & 1935 & 3259 & 42 & 1148.13 \\
179 & 1044 & 1517 & 56  & 455.00 & 
328 & 1962 & 3321 & 124 & 351.24 \\
188 & 1123 & 315283 & 562 & 1074.48 & 
330 & 1968 & 2983 & 100 & 1775.49 \\
192 & 1146 & 1841 & 75  & 1160.00 & 
331 & 1968 & 3277 & 44 & 612.88 \\
224 & 1386 & 2268 & 105  & 444.22 & 
335 & 1992 & 3250 & 47 & 1723.91 \\
226 & 1398 & 2225 & 36  & 948.15 & 
364 & 2154 & 3644 & 44 & 1692.04 \\
227 & 1398 & 2237 & 35  & 1002.38 & 
369 & 2190 & 3593 & 57 & 1218.3 \\
236 & 1470 & 2183 & 77  & 444.81 & 
384 & 2265 & 3765 & 41 & 930.11 \\
244 & 1527 & 2520 & 101  & 1536.84 & 
433 & 2502 & 4349 & 47 & 1002.86 \\
250 & 1563 & 2809 & 42  & 1190.49 & \_ & \_ & \_ & \_ & \_ \\
\midrule
Total & \_ & \_ & \_ & 22 & \_ & \_ & \_ & \_ & 21 \\
\bottomrule
\end{tabular}
}
\end{table}

\subsubsection{Comparison with a Concorde-based method}
Results in Table~\ref{tbl_concorde_first_1_50} and~\ref{tbl_concorde_first_51_100} show the comparison of our \ma with the Concorde-based method proposed in~\cite{DBLP:conf/ssci/MathiesonM20}. It is clear from the result that our proposed method outperforms the Concorde-based method not only in terms of solving the most number of instances but also in terms of solving the instances in less time (optimal values are shown in bold).
\begin{table}[h!]
\centering
\caption{Runtime comparison with~\cite{DBLP:conf/ssci/MathiesonM20} on the first 50 instances (labeled as `Concorde' since the authors use this software as the TSP-solver after the reduction). In the table, TW represents the treewidth. The approach presented in this work solved all of the first $100$ instances, while the previous method solved only half of them. The term ``Total" means the number of solved instances.}
\label{tbl_concorde_first_1_50}
\resizebox{0.99\textwidth}{!}{
\begin{tabular}{@{\extracolsep{4pt}}
p{0.8cm}cccp{1.8cm}c
p{0.8cm}cccp{1.8cm}c
@{}}
\toprule
\multirow{2}{*}{Inst.\#} & {\multirow{2}{*}{$\vert V \vert$}} & {\multirow{2}{*}{$\vert E \vert$}} & {\multirow{2}{*}{TW}} & This Approach & {\multirow{2}{*}{Concorde}} & \multirow{2}{*}{Inst.\#} & {\multirow{2}{*}{$\vert V \vert$}} & {\multirow{2}{*}{$\vert E \vert$}} & {\multirow{2}{*}{TW}} & {This Approach} & {\multirow{2}{*}{Concorde}}\\
\cmidrule{1-6}\cmidrule{7-12}
1 & 66 & 99 & 9 & \textbf{0.429} & 1.27 & 26 & 210 & 315 & 9 & \textbf{1.309} & \_\\
2 & 70 & 106 & 9 & 0.351 & \textbf{0.11} & 27 & 214 & 322 & 9 & \textbf{1.003} & 13.06\\
3 & 78 & 117 & 9 & \textbf{0.275} & 55.06 & 28 & 222 & 333 & 9 & \textbf{1.13} & \_\\
4 & 84 & 127 & 9 & 0.823 & \textbf{0.17} & 29 & 228 & 343 & 9 & \textbf{1.111} & \_\\
5 & 90 & 135 & 9 & \textbf{0.24} & 17.02 & 30 & 234 & 351 & 9 & \textbf{0.936} & \_\\
6 & 94 & 142 & 9 & 0.305 & \textbf{0.19} & 31 & 238 & 358 & 9 & 2.083 & \textbf{0.64} \\
7 & 102 & 153 & 9 & \textbf{0.367} & 16.27 & 32 & 246 & 369 & 9 & \textbf{1.133} & \_\\
8 & 108 & 163 & 9 & 1.232 & \textbf{0.17} & 33 & 252 & 379 & 9 & \textbf{1.734} & \_\\
9 & 114 & 171 & 9 & \textbf{0.458} & 81.97 & 34 & 258 & 387 & 9 & \textbf{1.623} & \_\\
10 & 118 & 178 & 9 & 0.553 & \textbf{0.36} & 35 & 262 & 394 & 9 & 4.292 & \textbf{0.34} \\
11 & 126 & 189 & 9 & \textbf{5.07} & 365 & 36 & 270 & 405 & 9 & \textbf{1.756} & \_\\
12 & 132 & 199 & 9 & \textbf{0.387} & 0.45 & 37 & 276 & 415 & 9 & \textbf{2.094} & \_\\
13 & 138 & 207 & 9 & 1.035 & \textbf{0.47} & 38 & 282 & 423 & 9 & \textbf{1.315} & \_\\
14 & 142 & 214 & 9 & \textbf{1.249} & 3.95 & 39 & 286 & 430 & 9 & 1.643 & \textbf{0.92} \\
15 & 150 & 225 & 9 & 0.708 & \textbf{0.61} & 40 & 294 & 441 & 9 & \textbf{1.673} & \_\\
16 & 156 & 235 & 9 & \textbf{1.344} & \_ & 41 & 300 & 451 & 9 & \textbf{4.539} & \_\\
17 & 162 & 243 & 9 & \textbf{0.777} & \_ & 42 & 306 & 459 & 9 & \textbf{2.56} & \_\\
18 & 166 & 250 & 9 & 0.971 & \textbf{0.53} & 43 & 310 & 466 & 9 & 1.571 & \textbf{0.77}\\
19 & 170 & 390 & 7 & \textbf{0.526} & \_ & 44 & 318 & 477 & 9 & \textbf{1.84} & \_\\
20 & 174 & 261 & 9 & \textbf{0.455} & \_ & 45 & 324 & 487 & 9 & \textbf{3.34} & \_\\
21 & 180 & 271 & 9 & 1.417 & \textbf{0.69} & 46 & 330 & 495 & 9 & \textbf{1.426} & \_\\
22 & 186 & 279 & 9 & \textbf{1.204} & 35.17 & 47 & 334 & 502 & 9 & \textbf{5.351} & 226.83\\
23 & 190 & 286 & 9 & 1.052 & \textbf{0.72} & 48 & 338 & 776 & 7 & \textbf{0.728} & \_\\
24 & 198 & 297 & 9 & \textbf{0.941} & \_ & 49 & 342 & 513 & 9 & \textbf{1.618} & \_\\
25 & 204 & 307 & 9 & \textbf{2.204} & \_ & 50 & 348 & 523 & 9 & \textbf{7.771} & \_\\
\bottomrule
Total & \_ & \_ & \_ & \textbf{25} & 19 & \_ & \_ & \_ & \_ & \textbf{25} & 6 \\
\bottomrule
\end{tabular}
}
\end{table}

\begin{table}[h!]
\centering
\caption{Runtime  comparison with ~\cite{DBLP:conf/ssci/MathiesonM20} on instances $51$ to $100$. In the table, TW represents the treewidth. The term ``Total" means the number of solved instances.}
\label{tbl_concorde_first_51_100}
\resizebox{0.99\textwidth}{!}{
\begin{tabular}{@{\extracolsep{4pt}}
p{0.6cm}cccp{1.8cm}c
p{0.6cm}cccp{1.8cm}c
@{}}
\toprule
\multirow{2}{*}{Inst.\#} & {\multirow{2}{*}{$\vert V \vert$}} & {\multirow{2}{*}{$\vert E \vert$}} & {\multirow{2}{*}{TW}} & {This Approach} & {\multirow{2}{*}{Concorde}} & \multirow{2}{*}{Inst.\#} & {\multirow{2}{*}{$\vert V \vert$}} & {\multirow{2}{*}{$\vert E \vert$}} & {\multirow{2}{*}{TW}} & {This Approach} & {\multirow{2}{*}{Concorde}}\\
\cmidrule{1-6}\cmidrule{7-12}
51 & 354 & 531 & 9 & \textbf{5.45} & \_ & 76 & 471 & 1161 & 7 & \textbf{0.65} & \_  \\
52 & 358 & 538 & 9 & 2.25 & \textbf{0.91} & 77 & 474 & 711 & 9 & \textbf{2.05} & \_  \\
53 & 366 & 549 & 9 & \textbf{2.93} & \_ & 78 & 478 & 718 & 9 & 5.52 & \textbf{2.03}  \\
54 & 372 & 559 & 9 & 3.01 & \textbf{1.03} & 79 & 480 & 57601 & 241 & 123.7 & \textbf{44.78}  \\
55 & 378 & 567 & 9 & \textbf{4.86} & \_ & 80 & 486 & 729 & 9 & \textbf{3.68} & \_  \\
56 & 382 & 574 & 9 & \textbf{2.8} & 10.5 & 81 & 492 & 739 & 9 & \textbf{13.02} & \_  \\
57 & 390 & 585 & 9 & \textbf{5.24} & \_ & 82 & 496 & 745 & 5 & \textbf{0.59} & 26.55  \\
58 & 396 & 595 & 9 & \textbf{3.01} & 5.7 & 83 & 498 & 747 & 9 & \textbf{2.38} & \_  \\
59 & 400 & 40001 & 201 & 75.7 & \textbf{27.48} & 84 & 500 & 62501 & 251 & 84.5 & \textbf{55.39}  \\
60 & 402 & 603 & 9 & \textbf{2.5} & \_ & 85 & 502 & 754 & 9 & \textbf{3.57} & \_  \\
61 & 406 & 610 & 9 & \textbf{8.46} & \_ & 86 & 503 & 1241 & 7 & \textbf{0.84} & \_  \\
62 & 408 & 936 & 7 & \textbf{0.68} & \_ & 87 & 507 & 1164 & 7 & \textbf{1.76} & \_  \\
63 & 414 & 621 & 9 & \textbf{4.31} & \_ & 88 & 507 & 1251 & 7 & \textbf{1.21} & \_  \\
64 & 416 & 625 & 6 & \textbf{0.36} & 229.48 & 89 & 510 & 765 & 9 & \textbf{8.79} & \_  \\
65 & 420 & 631 & 9 & \textbf{4.02} & \_ & 90 & 510 & 65026 & 256 & \textbf{61.7} & 70.62  \\
66 & 426 & 639 & 9 & \textbf{1.5} & \_ & 91 & 516 & 775 & 9 & \textbf{6.93} & \_  \\
67 & 430 & 646 & 9 & 3.783 & \textbf{1.19} & 92 & 522 & 783 & 9 & \textbf{7.01} & \_  \\
68 & 438 & 657 & 9 & \textbf{5.46} & \_ & 93 & 526 & 790 & 9 & 11.3 & \textbf{2.8}  \\
69 & 444 & 667 & 9 & \textbf{5.45} & \_ & 94 & 534 & 801 & 9 & \textbf{4.67} & \_  \\
70 & 450 & 675 & 9 & \textbf{6.54} & \_ & 95 & 540 & 811 & 9 & \textbf{11.28} & \_  \\
71 & 454 & 682 & 9 & \textbf{8.22} & 9.81 & 96 & 540 & 72901 & 271 & 281.1 & \textbf{62.67}  \\
72 & 460 & 52901 & 231 & 41.5 & \textbf{38.3} & 97 & 546 & 819 & 9 & \textbf{4.89} & \_  \\
73 & 462 & 693 & 9 & \textbf{3.3} & \_ & 98 & 546 & 825 & 36 & 83.3 & \textbf{14.69}  \\
74 & 462 & 756 & 34 & 391.02 & \textbf{7.88} & 99 & 550 & 826 & 9 & 9.02 & \textbf{7.8}  \\
75 & 468 & 703 & 9 & \textbf{11.96} & \_ & 100 & 558 & 837 & 9 & \textbf{6.47} & \_  \\
\midrule
Total & \_ & \_ & \_ & \textbf{25} & 10 & \_ & \_ & \_ & \_ & \textbf{25} & 9 \\
\bottomrule
\end{tabular}
}
\end{table}

Results in Table~\ref{tbl_hybridham_1} and Table~\ref{tbl_hybridham_2} show the comparison of our method with the algorithm proposed in~\cite{seeja2018hybridham} (hybridHAM). For each instance, the least runtime value is shown in bold. 
The hybridHAM method is only able to show the results for $75$ instances. Among those $75$ instances, they are able to find the \hc in only $13$ instances (our \ma found \hc in all $75$ instances). For all the remaining instances (among those $75-13 = 62$), hybridHAM is able to find the Hamiltonian Path (HP) only (shown with $``\_"$ in Table~\ref{tbl_hybridham_1} and~\ref{tbl_hybridham_2}). Clearly, the proposed \ma is the winner here in finding \hc for a more significant number of instances.

\begin{table}[h!]
\centering
\caption{Runtime comparison with HybridHAM~\cite{seeja2018hybridham} (results given in seconds). The $``\_"$ character in the Output column of the HybridHAM method is used when the algorithm was unable to find the Hamiltonian cycle (HC) for those instances (instead, they found a Hamiltonian Path in the time given in the `Runtime' column). The term ``Total" means the number of solved instances.}
\label{tbl_hybridham_1}
\resizebox{0.69\textwidth}{!}{
\begin{tabular}{@{\extracolsep{4pt}}cccccccc@{}}
\toprule
\multirow{2}{*}{Inst. \#} & {\multirow{2}{*}{$\vert V \vert$}} & \multirow{2}{*}{$\vert E \vert$} & \multirow{2}{*}{TW} & \multicolumn{2}{c}{HybridHAM} &\multicolumn{2}{c}{This Approach} 
\\
\cmidrule{5-6} \cmidrule{7-8}
 &&&& Output & Runtime  & Output & Runtime\\
\midrule
1 & 66 & 99 & 9 & \_ & 0.0625 & \textbf{HC} & \textbf{0.429} \\
2 & 70 & 106 & 9 & HC & \textbf{0.0469} & HC & 0.351 \\
3 & 78 & 117 & 9 & \_ & 0.0625 & \textbf{HC} & \textbf{0.275} \\
4 & 84 & 127 & 9  & \_ & 0.0625 & \textbf{HC} & \textbf{0.823} \\
5 & 90 & 135 & 9  & \_ & 0.0625 & \textbf{HC} & \textbf{0.24} \\
6 & 94 & 142 & 9  & HC & \textbf{0.0625} & HC & 0.305 \\
7 & 102 & 153 & 9  & \_ & 0.0781 & \textbf{HC} & \textbf{0.367} \\
8 & 108 & 162 & 9  & \_ & 0.0625 & \textbf{HC} & \textbf{1.232} \\
9 & 114 & 171 & 9  & \_ & 0.0938 & \textbf{HC} & \textbf{0.458} \\
11 & 126 & 189 & 9  & \_ & 0.1094 & \textbf{HC} & \textbf{5.07} \\
12 & 132 & 199 & 9  & \_ & 0.1094 & \textbf{HC} & \textbf{0.387} \\
14 & 142 & 214 & 9  & \_ & 0.0938 & \textbf{HC} & \textbf{1.249} \\
15 & 150 & 225 & 9  & \_ & 0.1250 & \textbf{HC} & \textbf{0.708} \\
16 & 156 & 235 & 9  & \_ & 0.0938 & \textbf{HC} & \textbf{1.344} \\
17 & 162 & 243 & 9  & \_ & 0.1875 & \textbf{HC} & \textbf{0.777} \\
18 & 166 & 250 & 9  & \_ & 0.1094 & \textbf{HC} & \textbf{0.971} \\
20 & 174 & 261 & 9  & \_ & 0.2344 & \textbf{HC} & \textbf{0.455} \\
21 & 180 & 271 & 9  & \_ & 0.1406 & \textbf{HC} & \textbf{1.417} \\
22 & 186 & 279 & 9  & \_ & 0.2188 & \textbf{HC} & \textbf{1.204} \\
23 & 190 & 286 & 9  & \_ & 0.1563 & \textbf{HC} & \textbf{1.052} \\
25 & 204 & 307 & 9  & \_ & 0.1719 & \textbf{HC} & \textbf{2.204} \\
26 & 210 & 315 & 9  & \_ & 0.3750 & \textbf{HC} & \textbf{1.309} \\
27 & 214 & 322 & 9  & \_ & 0.2031 & \textbf{HC} & \textbf{1.003} \\
29 & 228 & 343 & 9  & \_ & 0.2500 & \textbf{HC} & \textbf{1.111} \\
32 & 246 & 369 & 9  & \_ & 0.4063 & \textbf{HC} & \textbf{1.133} \\
33 & 252 & 379 & 9  & \_ & 0.3281 & \textbf{HC} & \textbf{1.734} \\
34 & 258 & 387 & 9  & \_ & 0.4688 & \textbf{HC} & \textbf{1.623} \\
35 & 262 & 394 & 9  & \_ & 0.5781 & \textbf{HC} & \textbf{4.292} \\
36 & 270 & 405 & 9  & \_ & 0.7031 & \textbf{HC} & \textbf{1.756} \\
37 & 276 & 415 & 9  & \_ & 0.3750  & \textbf{HC} & \textbf{2.094} \\
40 & 294 & 441 & 9  & \_ & 1.0469 & \textbf{HC} & \textbf{1.673} \\
41 & 300 & 451 & 9  & \_ & 0.4844 & \textbf{HC} & \textbf{4.539} \\
43 & 310 & 466 & 9  & \_ & 0.7031 & \textbf{HC} & \textbf{1.571} \\
44 & 312 & 477 & 9  & \_ & 0.9844 & \textbf{HC} & \textbf{1.84} \\
45 & 324 & 487 & 9  & \_ & 0.6094 & \textbf{HC} & \textbf{3.34} \\
50 & 348 & 523 & 9  & \_ & 0.9063 & \textbf{HC} & \textbf{7.771} \\
53 & 366 & 549 & 9 & \_ & 1.6875 & \textbf{HC} & \textbf{2.937} \\
54 & 372 & 559 & 9  & \_ & 0.9219 & \textbf{HC} & \textbf{3.01} \\
58 & 396 & 595 & 9  & \_ & 1.7031 & \textbf{HC} & \textbf{3.016} \\
59 & 400 & 40001 & 201  & HC & \textbf{0.2656} & HC & 75.7 \\
\midrule
Total & \_ & \_ & \_  & 3 & \_ & \textbf{40} & \_ \\
\bottomrule
\end{tabular}
}
\end{table}

\begin{table}[h!]
\centering
\caption{Runtime comparison with HybridHAM~\cite{seeja2018hybridham} (in seconds). The $``\_"$ in the Output column of the HybridHAM method indicates that their algorithm was unable to find the Hamiltonian cycle (HC) for those instances. Rather, they found the Hamiltonian Path (the path that does not start and end at the same vertex) in the time given in the Runtime column. The term ``Total" means the number of solved instances.}
\label{tbl_hybridham_2}
\resizebox{0.79\textwidth}{!}{
\begin{tabular}{@{\extracolsep{4pt}}
cccccccc@{}}
\toprule
\multirow{2}{*}{Inst. \#} & {\multirow{2}{*}{$\vert V \vert$}} & \multirow{2}{*}{$\vert E \vert$} & \multirow{2}{*}{TW} & \multicolumn{2}{c}{HybridHAM} &\multicolumn{2}{c}{This Approach} 
\\
\cmidrule{5-6} \cmidrule{7-8}
 &&&& Output & Runtime  & Output & Runtime\\
\midrule
64 & 416 & 625 & 6 & \_ & 1.0938 & \textbf{HC} & \textbf{0.368} \\
65 & 419 & 631 & 9  & \_ & 1.4375 & \textbf{HC} & \textbf{4.026} \\
68 & 438 & 657 & 9  & \_ & 2.9219 & \textbf{HC} & \textbf{5.467} \\
69 & 444 & 667 & 9  & \_ & 2.9063 & \textbf{HC} & \textbf{5.458} \\
72 & 460 & 52901 & 231  & HC & \textbf{0.4375} & HC & 41.5 \\
79 & 480 & 57601 & 241  & HC & \textbf{0.4844} & HC & 123.7 \\
82 & 496 & 745 & 5  & \_ & 1.7031 & \textbf{HC} & \textbf{0.597} \\
84 & 500 & 62501 & 251  & HC & \textbf{0.5313} & HC & 84.5 \\
90 & 510 & 65026 & 256  & HC & \textbf{0.5625} & HC & 61.7 \\
91 & 516 & 775 & 9  & \_ & 3.5156 & \textbf{HC} & \textbf{6.93} \\
95 & 540 & 811 & 9  & \_ & 3.0469 & \textbf{HC} & \textbf{11.283} \\
96 & 540 & 72901 & 271  & HC & \textbf{0.7031} & HC & 281.1 \\
99 & 550 & 826 & 9  & \_ & 5.6719 & \textbf{HC} & \textbf{9.026} \\
104 & 576 & 865 & 6  & \_ & 2.8594 & \textbf{HC} & \textbf{2.475} \\
118 & 636 & 955 & 9  & \_ & 6.8594 & \textbf{HC} & \textbf{1.734} \\
122 & 656 & 985 & 6  & \_ & 4.1563 & \textbf{HC} & \textbf{0.684} \\
124 & 660 & 991 & 9  & \_ & 8.2813 & \textbf{HC} & \textbf{1.672} \\
128 & 677 & 114583 & 339  & HC & \textbf{1.3594} & HC & 78.638 \\
134 & 724 & 131045 & 363  & HC & \textbf{1.5313} & HC & 198.901 \\
137 & 736 & 1105 & 6  & \_ & 12.9531 & \textbf{HC} & \textbf{2.789} \\
148 & 816 & 1225 & 6  & \_ & 7.9531 & \textbf{HC} & \textbf{10.517} \\
150 & 823 & 169333 & 412  & HC & \textbf{2.750} & HC & 453.815 \\
151 & 828 & 1243 & 9  & \_ & 11.5625 & \textbf{HC} & \textbf{4.937} \\
160 & 896 & 1345 & 6  & \_ & 14.00 & \textbf{HC} & \textbf{5.761} \\
162 & 909 & 206571 & 455  & HC & \textbf{4.0625} & HC & 66.752 \\
168 & 972 & 1459 & 9  & \_ & 33.0938 & \textbf{HC} & \textbf{4.325} \\
169 & 976 & 1465 & 6  & \_ & 28.6406 & \textbf{HC} & \textbf{1.577} \\
176 & 1020 & 1531 & 9  & \_ & 29.4375 & \textbf{HC} & \textbf{0.987} \\
182 & 1056 & 1585 & 6  & \_ & 22.6719 & \textbf{HC} & \textbf{18.191} \\
188 & 1123 & 315283 & 562  & HC & \textbf{9.6406} & HC & 1074.48 \\
190 & 1136 & 1705 & 6  & \_ & 14.4844 & \textbf{HC} & \textbf{0.546} \\
203 & 1216 & 1825 & 6  & \_ & 40.2813 & \textbf{HC} & \textbf{1.062} \\
211 & 1296 & 1945 & 6  & \_ & 40.1406 & \textbf{HC} & \textbf{1.265} \\
233 & 1456 & 2185 & 6  & \_ & 85.4688 & \textbf{HC} & \textbf{2.36}\\
246 & 1536 & 2305 & 6  & \_ & 111.843 & \textbf{HC} & \textbf{1.485} \\
\midrule
Total & \_ & \_ & \_  & 10 & \_ & \textbf{35} & \_ \\
\bottomrule
\end{tabular} 
}
\end{table}

\subsection{Rank-Based Comparison}
\begin{figure}[h!]
	\centering
	\includegraphics[scale=0.5] {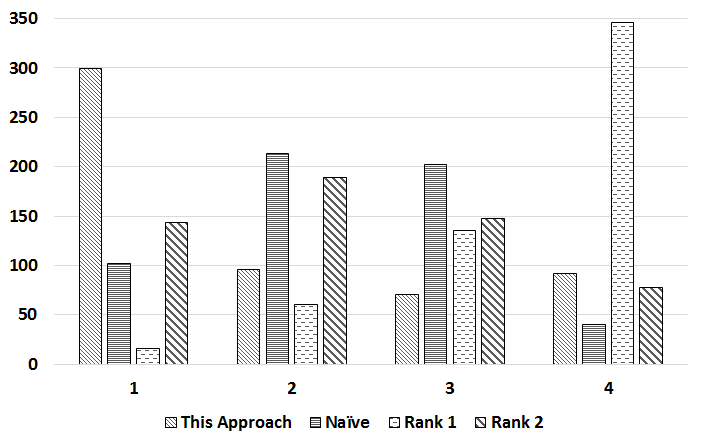}
	\caption{Histogram for the Rank comparison (ranks are given in the x-axis, and the y-axis shows the count of instances) of our proposed \ma with different baseline methods on set A instances (lower rank is better). }
	\label{fig_set_a_instances_hist_plot}
\end{figure}

Figure~\ref{fig_set_a_instances_hist_plot} shows the histogram with the quantitative distribution of the ranking results for instances of set A. 
We computed numerical ranking for the runtime using the \textit{rank()} function in Python. The instances that are solved with the lowest times are assigned a rank value $1$, while the highest runtime instances are assigned a rank value $4$. The runtime for instances having a rank greater than $1$ and less than $4$ are assigned accordingly. This process is carried out for our proposed \ma along with several baseline models. We then plot the histogram of those rankings (see Figure~\ref{fig_set_a_instances_hist_plot}) to evaluate the frequency of rank values in each of the $4$ bins.
We can see that the proposed method is better than the baselines because the majority rank values of the proposed method belong to the first bin (it means that the majority runtime values of solved instances for our \ma are on the lower side compared to the baseline models). Note that these results are shown only for those instances of set A that were solved by all methods.

\subsection{Discussion on the structure of Graphs}
\begin{figure}[h!]
	\centering
	\includegraphics[page=4, scale=0.6] {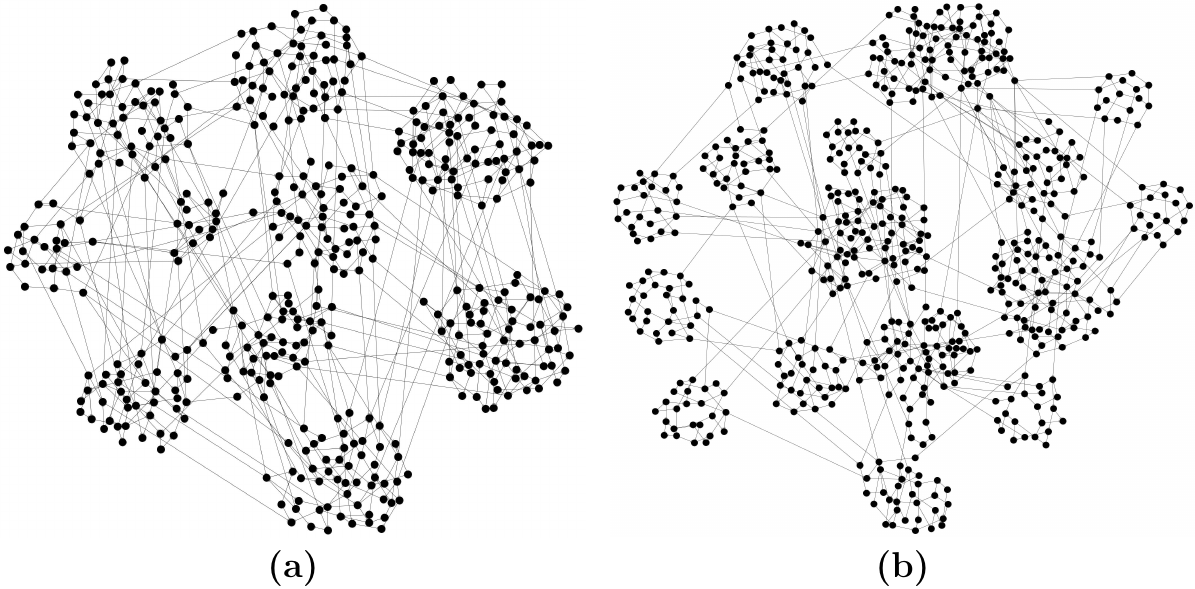}
	\caption{Different graph structures in \fhcpsc. (a) is instance $74$ (Treewidth~=~$34$), (b) is instance $109$ (Treewidth = $20$).}
	\label{fig_graph_struct_complex_vs_mpre_complex}
\end{figure}

To further investigate the behavior of our \maa, we analyze the structure of graphs. It is well known that it becomes difficult to find the \hc in a graph having higher treewidth as compared to a graph with lower treewidth. 
The structures of two graph instances, namely instance $74$ and instance $109$, are shown in Figure~\ref{fig_graph_struct_complex_vs_mpre_complex}. We can observe that both graphs have nodes divided into clusters. However, the number of edges between nodes of different clusters (inter-cluster edges) are in greater numbers for instance $74$ (Figure~\ref{fig_graph_struct_complex_vs_mpre_complex} (a)) as compared to instance $109$ (Figure~\ref{fig_graph_struct_complex_vs_mpre_complex} (b)). 
The complexities of these instances' structures can also be analyzed using the treewidth of these two graphs as the treewidth of graph $74$ is $34$, and the treewidth of graph $109$ is 20. This makes instance $74$ difficult in terms of finding the Hamiltonian cycle compared to instance $109$. 
Our \ma took less time to find the \hc in instance $109$ (because of smaller treewidth) as compared to instance $74$. This behavior shows the effectiveness of our algorithm as the proposed \ma is able to solve the complex instances, which are not solved by the existing methods from the literature (instance $74$ is not solved by any baseline method).

Similar behavior can be observed in the case of set F instances. The Rank $1$, Rank $2$, and Na\"{\i}ve Algorithm were unable to solve any instance from set F. This is due to the fact that instances in set F have greater treewidth (see Table~\ref{tbl_treewidth_set_f}). Since our sparsification method helps to reduce the treewidth of the graph, we were able to solve $43$ instances from set F, which were considered difficult for the baseline methods.

\begin{figure}[h!]
	\centering
	\includegraphics[page=6, scale=0.6] {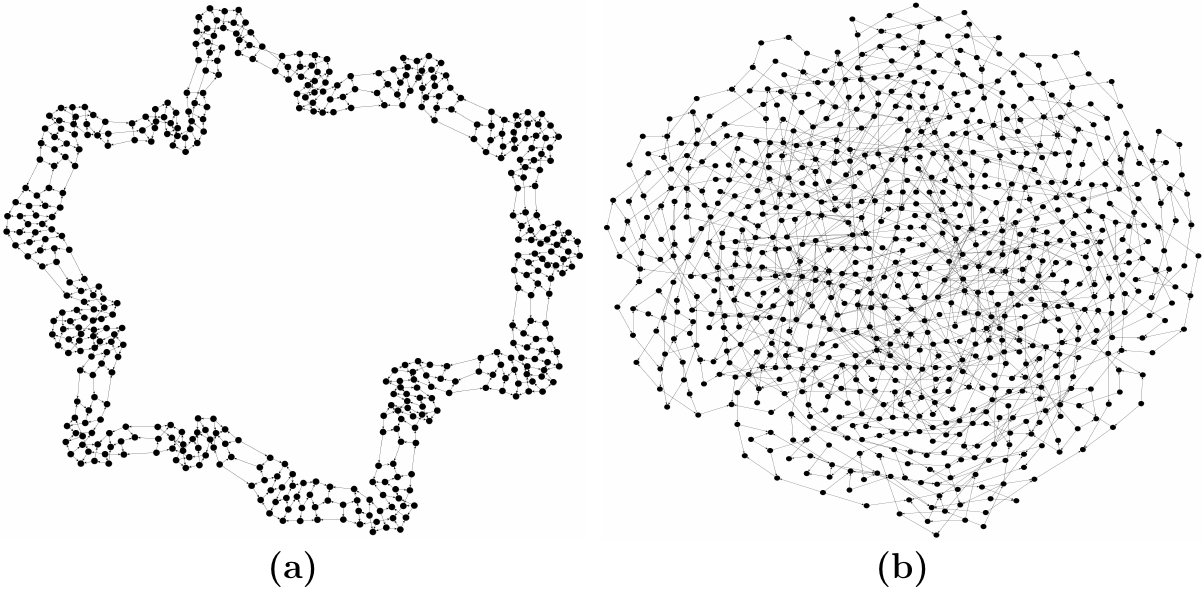}
	\caption{Example of different graph structures in \fhcpsc. (a) is instance $73$ (Treewidth = $9$), (b) is instance $149$ (Treewidth = $56$).}
	\label{fig_graph_struct_complex_vs_mpre_complex_73_74}
\end{figure}

Similarly, we can observe the graph structure of instance $73$ and $149$ in Figure~\ref{fig_graph_struct_complex_vs_mpre_complex_73_74} (a) and~\ref{fig_graph_struct_complex_vs_mpre_complex_73_74} (b), respectively. The structure of instance $73$ (Treewidth = $9$) is very simple as compared to instance $149$ (Treewidth = $56$), which is the reason why \ma was able to solve instance $73$ more quickly (in $3.3$ seconds) as compared to instance $149$ (in $51.03$ seconds). An interesting observation here is that regardless of the simple structure of instance $73$, no baseline method was able to find the Hamiltonian cycle in it. Since we are able to find a Hamiltonian cycle in simple (e.g., instance $73$) as well as more complex graph structures (e.g., instance $149$), we can conclude that our algorithm is more robust and is remarkably independent of the graph structure.
\subsection{Overall Summary of Results}
The average, standard deviation, maximum, and minimum value of the runtime for the instances solved by all methods in the first $100$, set A, set C, and set F instances are shown in Tables~\ref{tbl_overall_summary_first_100} and Table~\ref{tbl_overall_summary_set_C}. 
We have also provided the results for each instance separately for further analysis\footnote{\url{https://docs.google.com/spreadsheets/d/1tC0o6gyouNNNAOQret1bCXg_XyXKqRRv6uZrHbvX0m4/edit?usp=sharing}}.
\begin{table}[h!]
\centering
\caption{Runtime comparison (in sec.) for first $100$ and Set A instances.}
\label{tbl_overall_summary_first_100}
\begin{tabular}{@{\extracolsep{4pt}}p{0.6cm}
p{1.2cm}p{0.7cm}p{0.5cm}p{0.6cm}
p{1.2cm}p{0.7cm}p{0.5cm}p{0.6cm}
@{}}
\toprule
& \multicolumn{4}{c}{First $100$} & \multicolumn{4}{c}{Set A} 
\\
\cmidrule{2-5} \cmidrule{6-9}
 & This Approach & Na\"{\i}ve Algo.	 & Rank 1 & Rank 2 & This Approach & Na\"{\i}ve Algo.	 & Rank 1 & Rank 2 \\
\cmidrule{1-5} \cmidrule{6-9}
Avg. & \textbf{2.96} & 3.36 & 5.9 & 5.51 & \textbf{69.27} & 82.02 & 93.78 & 71.31 \\
Median & \textbf{1.74} & 2.02 & 4.27 & 4.07 & \textbf{23.81} & 40.17 & 45.92 & 37.18 \\
Max. & \textbf{13.02} & 20.35 & 27.5 & 23.2 & 513.06 & 560.8 & 554.9 & \textbf{506.7} \\
Min. & 0.24 & \textbf{0.07} & 0.17 & 0.17 & 0.24 & \textbf{0.07} & 0.171 & 0.171 \\
\bottomrule
\end{tabular}
\end{table}

\begin{table}[h!]
\centering
\caption{Runtime comparison (in sec.) for Set C and Set F instances.}
\label{tbl_overall_summary_set_C}
\begin{tabular}{@{\extracolsep{4pt}}p{0.4cm}p{1.2cm}p{0.9cm}cccc@{}}
\toprule
& \multicolumn{5}{c}{Set C} & \multicolumn{1}{c}{Set F} \\
\cmidrule{2-6} \cmidrule{7-7}
 & This Approach & Na\"{\i}ve Algo.	 & \multirow{2}{*}{Rank 1} & \multirow{2}{*}{Rank 2} & \multirow{2}{*}{Concorde} & \multirow{2}{*}{This Approach} \\
\cmidrule{1-1} \cmidrule{2-6} \cmidrule{7-7}
Avg. & 21.04 & 133.77 & 248.72 & 52.84  & \textbf{15.71} & 777.50\\
Med. & 26.44 & 29.17 & 244.50 & 29.84  &  \textbf{14.71} & 739.28  \\
Max. & \textbf{30.05} & 476.07 & 504.85 & 150.99  &  30.97& 1775.4  \\
Min. & 1.225 & \textbf{0.671} & 1.031 & 0.703 & 2.47 & 3.3   \\
\bottomrule
\end{tabular}
\end{table}

\subsection{Association of the running time with the number of vertices and edges}
\begin{figure}[h!]
\centering
\includegraphics{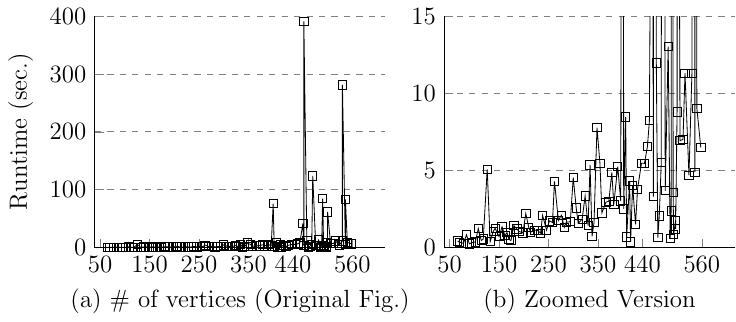}
\caption{Runtime analysis for the first $100$ instances (with increasing number of vertices) using our \maa. Figure (a) shows the original plot for the runtime, while Figure (b) is the zoomed version of Figure (a).}
\label{fig_runtime_analysis_first_100}
\end{figure}

\begin{figure}[h!]
\centering
\includegraphics{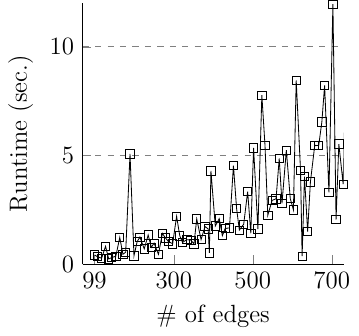}
\caption{Runtime analysis for an increasing number of edges using our \maa.}
\label{fig_runtime_analysis_edges_first_100}
\end{figure}

Figure~\ref{fig_runtime_analysis_first_100} shows the effect of the number of vertices on the runtime for our \ma (on the first $100$ instances). We can observe that as we increase the number of vertices, the runtime slowly increases apart from a few outliers (a kind of linear increase). Figure~\ref{fig_runtime_analysis_edges_first_100} shows the effect of increasing the number of edges on the runtime. 
In Figure~\ref{fig_runtime_analysis_edges_first_100}, the increase in runtime with an increasing number of edges is slower as compared to the increase in runtime with an increasing number of vertices in Figure~\ref{fig_runtime_analysis_first_100}.


\section{Conclusion}\label{section_conclusion}

We propose an efficient memetic algorithm that uses the sparsification and augmentation approach along with the \rai and \lkh local search heuristic and the Na\"{\i}ve Algorithm's power to find the Hamiltonian cycle in the given graph. We highlighted that this dynamical approach reduces the time to find the \hc in a given graph.

We showed through results that our proposed \ma is simple and can be applied to any type of graph regardless of its structure and complexity.
We outperform the existing state-of-the-art methods both in terms of runtime and also in terms of finding the \hc in most number of instances. As a possible direction for the future, we would like to explore if other methods to sparsify the graph can be proposed so that the difficult instances can be solved.

\section{Related Work} \label{sec_related_work}
In this section, we provide information about some other approaches that exist for the Hamiltonian cycle problem.

First, we note that different types of methods have been proposed in the literature to solve the Hamiltonian cycle problem (\hcpp). Some methods are efficient in terms of runtime but expensive in space complexity. Other algorithms take less space but are computationally very time-consuming. Some methods are proposed for special structures of the graphs only (i.e., graphs with low treewidth) but failed to find \hc in the instances with complex structures. We will now discuss each type of solution separately.

\subsection{Algorithms with time and/or space complexity bounds}
For many algorithmic problems on graphs of treewidth $t$, a commonly used dynamic programming approach can be used to design an algorithm with time and space complexity $2^{O(t)} \cdot n^{O(1)}$~\cite{cygan2015parameterized,cygan2011solving,belbasi2019space}.
A lot of efforts have been made previously to further reduce the time and space complexities. 
Some algorithms (e.g., Held-Karp algorithm, which is based on dynamic programming ($O(n^{2} 2^{n})$)~\cite{held1962dynamic}) are cheap in terms of time complexity. However,~\cite{held1962dynamic} is expensive in terms of space complexity. Other methods (e.g., depth-first search based algorithms~\cite{rubin1974search,cheeseman1991really,vandegriend1998gn,van2018predictive}) are efficient in terms of space complexity. However, they are expensive in terms of time complexity (e.g., the theoretical worst-case time is $O(|V|!)$). V. Bagaria et al. in~\cite{bagaria2020hidden} introduced an approach to solving the \hcp using linear programming. They first reduce the \hcp to \ts problem (\tspp) and then apply a simple linear programming (LP) relaxation (the fractional $2$-factor ``F2F" LP) to solve the \hcp problem in the graph. 
One of the major problems with the standard dynamic programming approach is that it tends to have prohibitively large space complexity~\cite{belbasi2019space}. 
It has been proved theoretically in~\cite{drucker2016exponential} that give graphs of bounded treewidth, reducing the space complexity for the dynamic programming-based algorithm is not possible for many \npp-Hard problems. They show that under some feasible assumptions, it is not possible to design a dynamic programming-based algorithm, which takes $2^{O(t)} \cdot n^{O(1)}$ time and space for solving the 3-Coloring or Independent Set problem.
Authors in~\cite{belbasi2019space} proposed a dynamic programming-based algorithm, which uses polynomial space with a slight increase in the runtime (compared to standard dynamic algorithm-based methods) to solve \hcp and \tspp. 
Their proposed algorithm finds a Hamiltonian cycle in the graph in $O((4w)^{d} \ n M (n \ log \ n))$ time and $O(dn \ log \ n)$ space, where $w$ represents the width, $d$ represents the depth of the tree decomposition, and  $M(r)$ is the time complexity to multiply two $r$ bits integers.

\subsection{Methods based on graph structure}
A lot of recent research on the Hamiltonian cycle problem (\hcpp) is focused on solving the specific structure of the graphs (e.g., cactus, tree, quasi-threshold graph, etc.). 
A method to the \hcp is proposed in~\cite{ziobro2019finding} in which the authors argue that the graph parameter treewidth has a great impact on the worst-case performance of graph algorithms. It is known that many computational problems, including \npp-complete, can be solved efficiently for graphs of bounded treewidth. Eppstein in~\cite{eppstein2007traveling} proposes that \hcp can be solved in a graph of degree $\leq 3$ in time complexity equals to $O(2^{\frac{n}{3}}) \approx 1.26^{n}$ by taking linear space. His algorithm 
\cite{eppstein2007traveling} is also capable of computing an optimum solution for \tsp on such graphs. A recent algorithm working with degree-three graphs is proposed in~\cite{liskiewicz2014new} and~\cite{xiao2016exact}, which computes the tour in $O(1.2553^{n})$ and $O^{*}(2^{\frac{3n}{10}}) \approx 1.2312^{n}$ respectively, where $O^{*}(.)$ suppresses the polynomial factors.

\subsection{Methods that do not rely on graph structure}
There is not much work done in designing a general approach to solving the \hcp efficiently. Authors in~\cite{maretic2015heuristics} propose a modification of the ant colony optimization (\textsc{aco}) algorithm to solve the Hamiltonian completion problem (finding the least number of extra edges to make the graph Hamiltonian). K. Puljic and R. Manger in ~\cite{puljic2020evolutionary} use evolutionary algorithms (containing multiple crossover and mutation operators) to solve the Hamiltonian completion problem.

\subsection{Efficient methods that use local search}
Like us, others take advantage that the \hcp can be reduced to \tspp. Smaller \tsp instances can be formulated as mixed-integer linear programming (\textsc{MILP}) instances. These \textsc{MILP} instances can be solved using methods such as branch-and-cut, and branch-and-bound~\cite{korte2012combinatorial} (one popular method to solve the \tsp with a branch-and-cut technique is by using Concorde~\cite{Concorde}). {\em local-search heuristics} (e.g. $2$-opt, $3$-opt~\cite{papadimitriou1998combinatorial}) or more complex and powerful meta-heuristics 
(e.g memetic algorithms~\cite{DBLP:books/sp/19/MoscatoM19}, genetic algorithms~\cite{michalewicz1996genetic}) can be used to solve the \tsp~\cite{DBLP:journals/heuristics/BuriolFM04,DBLP:journals/memetic/EremeevK20}. However, due to the intractability of the \tspp, it becomes very difficult to get the exact solution using these approaches as the size of the graph grows (i.e., only an approximate solution is possible in the majority of the scenarios). 

An efficient local search method for the \tspp, called Lin-Kernighan heuristic (\lkhh), is proposed in~\cite{korte2012combinatorial}. 
The problem with $2$-opt and $3$-opt is that they usually converge to a nearby local optimum. Therefore, they are not very accurate. Memetic algorithm ($\mathsf{MAs}$) uses different local search methods (e.g., RAI, \lkhh), and they have specialized recombination methods to reduce the likelihood of the premature convergence~\cite{DBLP:journals/compsys/MerzF02}. These combined approaches then tend to be better than $2$-opt and $3$-opt techniques. 

Recently, a lot of work has focused on using genetic and evolutionary algorithms to solve the \tsp~\cite{ahmed2014improved,contreras2015automatic,puljic2020evolutionary}. This is because these techniques allow flexible control over the trade-off between computational time and accuracy (i.e., accuracy $\propto$ time).

\subsection{Deep Learning based methods}
Deep learning methods can be trained for combinatorial optimization problems (called \textit{Learning to Optimise}~\cite{li2017learning}) in order to predict their solution. Learning to Optimise field is divided into two categories namely \textit{supervised}~\cite{vinyals2015pointer} and \textit{reinforcement learning} (\textsc{rl})~\cite{deudon2018learning}. Several methods have been proposed in the literature, which use supervised and \textsc{rl} based methods to solve classical combinatorial optimization problems like \tsp~\cite{vinyals2015pointer}, Knapsack~\cite{bello2016neural}, and Vehicle Routing problems~\cite{nazari2018reinforcement}.

\subsection{Other unconventional computing paradigms}
Several authors used unconventional computing paradigms to solve the Hamiltonian cycle problem~\cite{mahasinghe2019solving,sleegers2020plant}. Different methods have been proposed in the literature to find the Hamiltonian cycle in optical computing~\cite{oltean2008solving}, DNA computing~\cite{lee1999dna}, and quantum computing~\cite{mahasinghe2019solving,hao2001quantum,lucas2014ising}. Although quantum computing is popular among the research community for solving many hard problems (as compared to the other mentioned methods)~\cite{cao2013quantum,harrow2009quantum,mahasinghe2016efficient}, it is still not possible to solve the \npp-complete problems in polynomial time using quantum computers.

\subsection{Known solvers for \tsp}

\subsubsection{The \lkh heuristic}
A highly successful, varied, and enhanced implementation of the famous Lin-Kernighan heuristic~\cite{HELSGAUN2000106} is known for its success of having discovered the best-known solutions for many benchmark examples, including the World \tsp Challenge~\footnote{\url{http://www.math.uwaterloo.ca/tsp/world/}}. It is now used in many algorithms around the world to compute near-optimal solutions to large instances of \tspp.
LKH is \textit{non-deterministic heuristic} because it relies on randomized methods to obtain \hcc. 
The source code for \lkh heuristic is available online~\footnote{\url{http://www.math.uwaterloo.ca/tsp/concorde/downloads/downloads.htm}}.

\subsubsection{Concorde}
Concorde~\cite{Concorde}, a well-known exact \tsp solver, uses a variety of algorithmic strategies to obtain and verify an optimal solution for the \tspp. It is generally assumed that one drawback of Concorde is that the verification process takes much longer than obtaining the actual solution. However, studies like the one in~\cite{DBLP:journals/heuristics/MuDHS18} now give a better and more precise account of their scalability and performance, so we recommend these relatively recent references (and work cited there) as an introduction to the topic. The source code of Concorde is available online~\footnote{\url{http://www.math.uwaterloo.ca/tsp/concorde.html}}.

\section{Acknowledgements}
We thank Prof. Regina Berretta and Dr. Mohammad Nazmul Haque from the College of Engineering Science and Environment at the University of Newcastle, Australia, Dr. Markus Wagner from the School of Computer Science at The University of Adelaide, Australia, and Prof. Luciana Buriol of the Institute of Informatics, Federal University of Rio Grande do Sul (UFRGS), Brazil for their thoughtful comments that helped us to improve the earlier versions of the manuscript. Special thanks also go to Dr. Luke Mathieson from the School of Computer Science, University of Technology Sydney. 

\bibliographystyle{elsarticle-harv}
\bibliography{references}

\begin{thebibliography}{89}
\expandafter\ifx\csname natexlab\endcsname\relax\def\natexlab#1{#1}\fi
\providecommand{\url}[1]{\texttt{#1}}
\providecommand{\href}[2]{#2}
\providecommand{\path}[1]{#1}
\providecommand{\DOIprefix}{doi:}
\providecommand{\ArXivprefix}{arXiv:}
\providecommand{\URLprefix}{URL: }
\providecommand{\Pubmedprefix}{pmid:}
\providecommand{\doi}[1]{\href{http://dx.doi.org/#1}{\path{#1}}}
\providecommand{\Pubmed}[1]{\href{pmid:#1}{\path{#1}}}
\providecommand{\bibinfo}[2]{#2}
\ifx\xfnm\relax \def\xfnm[#1]{\unskip,\space#1}\fi
\bibitem[{Ahammed and Moscato(2011)}]{DBLP:conf/evoW/AhammedM11}
\bibinfo{author}{Ahammed, F.}, \bibinfo{author}{Moscato, P.},
  \bibinfo{year}{2011}.
\newblock \bibinfo{title}{Evolving l-systems as an intelligent design approach
  to find classes of difficult-to-solve traveling salesman problem instances},
  in: \bibinfo{booktitle}{Applications of Evolutionary Computation Part {I}},
  pp. \bibinfo{pages}{1--11}.
\bibitem[{Ahmed(2014)}]{ahmed2014improved}
\bibinfo{author}{Ahmed, Z.H.}, \bibinfo{year}{2014}.
\newblock \bibinfo{title}{Improved genetic algorithms for the travelling
  salesman problem}.
\newblock \bibinfo{journal}{International Journal of Process Management and
  Benchmarking} \bibinfo{volume}{4}, \bibinfo{pages}{109--124}.
\bibitem[{Applegate et~al.(2006a)Applegate, Bixby, Chvatal and Cook}]{Concorde}
\bibinfo{author}{Applegate, D.}, \bibinfo{author}{Bixby, R.},
  \bibinfo{author}{Chvatal, V.}, \bibinfo{author}{Cook, W.},
  \bibinfo{year}{2006}a.
\newblock \bibinfo{title}{Concorde tsp solver} \URLprefix
  \url{http://www.math.uwaterloo.ca/tsp/concorde/}.
\bibitem[{Applegate et~al.(2006b)Applegate, Bixby, Chvatal and
  Cook}]{applegate2006traveling}
\bibinfo{author}{Applegate, D.L.}, \bibinfo{author}{Bixby, R.E.},
  \bibinfo{author}{Chvatal, V.}, \bibinfo{author}{Cook, W.J.},
  \bibinfo{year}{2006}b.
\newblock \bibinfo{title}{The traveling salesman problem: a computational
  study}.
\bibitem[{Bagaria et~al.(2020)Bagaria, Ding, Tse, Wu and
  Xu}]{bagaria2020hidden}
\bibinfo{author}{Bagaria, V.}, \bibinfo{author}{Ding, J.},
  \bibinfo{author}{Tse, D.}, \bibinfo{author}{Wu, Y.}, \bibinfo{author}{Xu,
  J.}, \bibinfo{year}{2020}.
\newblock \bibinfo{title}{Hidden hamiltonian cycle recovery via linear
  programming}.
\newblock \bibinfo{journal}{Operations research} \bibinfo{volume}{68},
  \bibinfo{pages}{53--70}.
\bibitem[{Baniasadi et~al.(2014)Baniasadi, Ejov, Filar, Haythorpe and
  Rossomakhine}]{baniasadi2014deterministic}
\bibinfo{author}{Baniasadi, P.}, \bibinfo{author}{Ejov, V.},
  \bibinfo{author}{Filar, J.A.}, \bibinfo{author}{Haythorpe, M.},
  \bibinfo{author}{Rossomakhine, S.}, \bibinfo{year}{2014}.
\newblock \bibinfo{title}{Deterministic ``snakes and ladders” heuristic for
  the hamiltonian cycle problem}.
\newblock \bibinfo{journal}{Mathematical Programming Computation}
  \bibinfo{volume}{6}, \bibinfo{pages}{55--75}.
\bibitem[{Belbasi and F{\"u}rer(2019)}]{belbasi2019space}
\bibinfo{author}{Belbasi, M.}, \bibinfo{author}{F{\"u}rer, M.},
  \bibinfo{year}{2019}.
\newblock \bibinfo{title}{A space-efficient parameterized algorithm for the
  hamiltonian cycle problem by dynamic algebraization}, in:
  \bibinfo{booktitle}{International Computer Science Symposium in Russia}, pp.
  \bibinfo{pages}{38--49}.
\bibitem[{Bello et~al.(2017)Bello, Pham, Le, Norouzi and
  Bengio}]{bello2016neural}
\bibinfo{author}{Bello, I.}, \bibinfo{author}{Pham, H.}, \bibinfo{author}{Le,
  Q.V.}, \bibinfo{author}{Norouzi, M.}, \bibinfo{author}{Bengio, S.},
  \bibinfo{year}{2017}.
\newblock \bibinfo{title}{Neural combinatorial optimization with reinforcement
  learning}, in: \bibinfo{booktitle}{Workshop track - International Conference
  on Learning Representation}, pp. \bibinfo{pages}{1--5}.
\bibitem[{Bentley(1990)}]{bentley1990experiments}
\bibinfo{author}{Bentley, J.}, \bibinfo{year}{1990}.
\newblock \bibinfo{title}{Experiments on traveling salesman heuristics}, in:
  \bibinfo{booktitle}{Symposium on Discrete algorithms}, pp.
  \bibinfo{pages}{91--99}.
\bibitem[{Berretta et~al.(2004)Berretta, Cotta and Moscato}]{Berretta2004}
\bibinfo{author}{Berretta, R.}, \bibinfo{author}{Cotta, C.},
  \bibinfo{author}{Moscato, P.}, \bibinfo{year}{2004}.
\newblock \bibinfo{title}{Enhancing the Performance of Memetic Algorithms by
  Using a Matching-Based Recombination Algorithm}.
\newblock pp. \bibinfo{pages}{65--90}.
\bibitem[{Berretta and Rodrigues(2004)}]{BERRETTA200467}
\bibinfo{author}{Berretta, R.}, \bibinfo{author}{Rodrigues, L.F.},
  \bibinfo{year}{2004}.
\newblock \bibinfo{title}{A memetic algorithm for a multistage capacitated
  lot-sizing problem}.
\newblock \bibinfo{journal}{International Journal of Production Economics}
  \bibinfo{volume}{87}, \bibinfo{pages}{67--81}.
\bibitem[{Bodlaender et~al.(2015)Bodlaender, Cygan, Kratsch and
  Nederlof}]{bodlaender2015deterministic}
\bibinfo{author}{Bodlaender, H.L.}, \bibinfo{author}{Cygan, M.},
  \bibinfo{author}{Kratsch, S.}, \bibinfo{author}{Nederlof, J.},
  \bibinfo{year}{2015}.
\newblock \bibinfo{title}{Deterministic single exponential time algorithms for
  connectivity problems parameterized by treewidth}.
\newblock \bibinfo{journal}{Information and Computation} \bibinfo{volume}{243},
  \bibinfo{pages}{86--111}.
\bibitem[{Buriol et~al.(2004)Buriol, Fran{\c{c}}a and
  Moscato}]{DBLP:journals/heuristics/BuriolFM04}
\bibinfo{author}{Buriol, L.S.}, \bibinfo{author}{Fran{\c{c}}a, P.M.},
  \bibinfo{author}{Moscato, P.}, \bibinfo{year}{2004}.
\newblock \bibinfo{title}{A new memetic algorithm for the asymmetric traveling
  salesman problem}.
\newblock \bibinfo{journal}{Journal of Heuristics} \bibinfo{volume}{10},
  \bibinfo{pages}{483--506}.
\bibitem[{Burke et~al.(2003)Burke, Hart, Kendall, Newall, Ross and
  Shulenburg}]{burke2emerging}
\bibinfo{author}{Burke, E.}, \bibinfo{author}{Hart, E.},
  \bibinfo{author}{Kendall, G.}, \bibinfo{author}{Newall, J.},
  \bibinfo{author}{Ross, P.}, \bibinfo{author}{Shulenburg, S.},
  \bibinfo{year}{2003}.
\newblock \bibinfo{title}{An emerging direction in modern search technology}.
\newblock \bibinfo{journal}{Handbook of Metaheuristics} \bibinfo{volume}{2},
  \bibinfo{pages}{457474}.
\bibitem[{Cao et~al.(2013)Cao, Papageorgiou, Petras, Traub and
  Kais}]{cao2013quantum}
\bibinfo{author}{Cao, Y.}, \bibinfo{author}{Papageorgiou, A.},
  \bibinfo{author}{Petras, I.}, \bibinfo{author}{Traub, J.},
  \bibinfo{author}{Kais, S.}, \bibinfo{year}{2013}.
\newblock \bibinfo{title}{Quantum algorithm and circuit design solving the
  poisson equation}.
\newblock \bibinfo{journal}{New Journal of Physics} \bibinfo{volume}{15},
  \bibinfo{pages}{013021}.
\bibitem[{Cheeseman et~al.(1991)Cheeseman, Kanefsky and
  Taylor}]{cheeseman1991really}
\bibinfo{author}{Cheeseman, P.C.}, \bibinfo{author}{Kanefsky, B.},
  \bibinfo{author}{Taylor, W.M.}, \bibinfo{year}{1991}.
\newblock \bibinfo{title}{Where the really hard problems are.}, in:
  \bibinfo{booktitle}{International Joint Conference on Artificial
  Intelligence}, pp. \bibinfo{pages}{331--337}.
\bibitem[{Cinar et~al.(2020)Cinar, Korkmaz and Kiran}]{cinar2020discrete}
\bibinfo{author}{Cinar, A.C.}, \bibinfo{author}{Korkmaz, S.},
  \bibinfo{author}{Kiran, M.S.}, \bibinfo{year}{2020}.
\newblock \bibinfo{title}{A discrete tree-seed algorithm for solving symmetric
  traveling salesman problem}.
\newblock \bibinfo{journal}{Engineering Science and Technology, an
  International Journal} \bibinfo{volume}{23}, \bibinfo{pages}{879--890}.
\bibitem[{Contreras-Bolton and Parada(2015)}]{contreras2015automatic}
\bibinfo{author}{Contreras-Bolton, C.}, \bibinfo{author}{Parada, V.},
  \bibinfo{year}{2015}.
\newblock \bibinfo{title}{Automatic combination of operators in a genetic
  algorithm to solve the traveling salesman problem}.
\newblock \bibinfo{journal}{PloS one} \bibinfo{volume}{10},
  \bibinfo{pages}{e0137724}.
\bibitem[{Cygan et~al.(2015)Cygan, Fomin, Kowalik, Lokshtanov, Marx, Pilipczuk,
  Pilipczuk and Saurabh}]{cygan2015parameterized}
\bibinfo{author}{Cygan, M.}, \bibinfo{author}{Fomin, F.},
  \bibinfo{author}{Kowalik, {\L}.}, \bibinfo{author}{Lokshtanov, D.},
  \bibinfo{author}{Marx, D.}, \bibinfo{author}{Pilipczuk, M.},
  \bibinfo{author}{Pilipczuk, M.}, \bibinfo{author}{Saurabh, S.},
  \bibinfo{year}{2015}.
\newblock \bibinfo{title}{Parameterized algorithms}.
  volume~\bibinfo{volume}{4}.
\bibitem[{Cygan et~al.(2018)Cygan, Kratsch and Nederlof}]{cygan2018fast}
\bibinfo{author}{Cygan, M.}, \bibinfo{author}{Kratsch, S.},
  \bibinfo{author}{Nederlof, J.}, \bibinfo{year}{2018}.
\newblock \bibinfo{title}{Fast hamiltonicity checking via bases of perfect
  matchings}.
\newblock \bibinfo{journal}{Journal of the ACM} \bibinfo{volume}{65},
  \bibinfo{pages}{1--46}.
\bibitem[{Cygan et~al.(2011)Cygan, Nederlof, Pilipczuk, Pilipczuk, van Rooij
  and Wojtaszczyk}]{cygan2011solving}
\bibinfo{author}{Cygan, M.}, \bibinfo{author}{Nederlof, J.},
  \bibinfo{author}{Pilipczuk, M.}, \bibinfo{author}{Pilipczuk, M.},
  \bibinfo{author}{van Rooij, J.M.}, \bibinfo{author}{Wojtaszczyk, J.O.},
  \bibinfo{year}{2011}.
\newblock \bibinfo{title}{Solving connectivity problems parameterized by
  treewidth in single exponential time}, in: \bibinfo{booktitle}{Annual
  Symposium on Foundations of Computer Science}, pp. \bibinfo{pages}{150--159}.
\bibitem[{Deudon et~al.(2018)Deudon, Cournut, Lacoste, Adulyasak and
  Rousseau}]{deudon2018learning}
\bibinfo{author}{Deudon, M.}, \bibinfo{author}{Cournut, P.},
  \bibinfo{author}{Lacoste, A.}, \bibinfo{author}{Adulyasak, Y.},
  \bibinfo{author}{Rousseau, L.M.}, \bibinfo{year}{2018}.
\newblock \bibinfo{title}{Learning heuristics for the tsp by policy gradient},
  in: \bibinfo{booktitle}{International Conference on the Integration of
  Constraint Programming, Artificial Intelligence, and Operations Research},
  pp. \bibinfo{pages}{170--181}.
\bibitem[{Dirac(1952)}]{dirac1952some}
\bibinfo{author}{Dirac, G.A.}, \bibinfo{year}{1952}.
\newblock \bibinfo{title}{Some theorems on abstract graphs}.
\newblock \bibinfo{journal}{Proceedings of the London Mathematical Society}
  \bibinfo{volume}{3}, \bibinfo{pages}{69--81}.
\bibitem[{Dogrus{\"o}z and Krishnamoorthy(1995)}]{dogrusoz1995hamiltonian}
\bibinfo{author}{Dogrus{\"o}z, U.}, \bibinfo{author}{Krishnamoorthy, M.},
  \bibinfo{year}{1995}.
\newblock \bibinfo{title}{Hamiltonian cycle problem for triangle graphs} .
\bibitem[{Drucker et~al.(2016)Drucker, Nederlof and
  Santhanam}]{drucker2016exponential}
\bibinfo{author}{Drucker, A.}, \bibinfo{author}{Nederlof, J.},
  \bibinfo{author}{Santhanam, R.}, \bibinfo{year}{2016}.
\newblock \bibinfo{title}{Exponential time paradigms through the polynomial
  time lens}, in: \bibinfo{booktitle}{Annual European Symposium on Algorithms}.
\bibitem[{Eppstein(2007)}]{eppstein2007traveling}
\bibinfo{author}{Eppstein, D.}, \bibinfo{year}{2007}.
\newblock \bibinfo{title}{The traveling salesman problem for cubic graphs}.
\newblock \bibinfo{journal}{Journal of Graph Algorithms and Applications}
  \bibinfo{volume}{11}, \bibinfo{pages}{61--81}.
\bibitem[{Eremeev and Kovalenko(2020)}]{DBLP:journals/memetic/EremeevK20}
\bibinfo{author}{Eremeev, A.V.}, \bibinfo{author}{Kovalenko, Y.V.},
  \bibinfo{year}{2020}.
\newblock \bibinfo{title}{A memetic algorithm with optimal recombination for
  the asymmetric travelling salesman problem}.
\newblock \bibinfo{journal}{Memetic Computing} \bibinfo{volume}{12},
  \bibinfo{pages}{23--36}.
\bibitem[{{Escobar} et~al.(2016){Escobar}, {Hidalgo}, {Inostroza-Ponta},
  {Marín}, {Rosas} and {Dorn}}]{7836019}
\bibinfo{author}{{Escobar}, I.}, \bibinfo{author}{{Hidalgo}, N.},
  \bibinfo{author}{{Inostroza-Ponta}, M.}, \bibinfo{author}{{Marín}, M.},
  \bibinfo{author}{{Rosas}, E.}, \bibinfo{author}{{Dorn}, M.},
  \bibinfo{year}{2016}.
\newblock \bibinfo{title}{Evaluation of a combined energy fitness function for
  a distributed memetic algorithm to tackle the 3d protein structure prediction
  problem}, in: \bibinfo{booktitle}{International Conference of the Chilean
  Computer Science Society}, pp. \bibinfo{pages}{1--10}.
\bibitem[{Franca et~al.(2006)Franca, Tin~Jr and Buriol}]{franca2006genetic}
\bibinfo{author}{Franca, P.M.}, \bibinfo{author}{Tin~Jr, G.},
  \bibinfo{author}{Buriol, L.}, \bibinfo{year}{2006}.
\newblock \bibinfo{title}{Genetic algorithms for the no-wait flowshop
  sequencing problem with time restrictions}.
\newblock \bibinfo{journal}{International Journal of Production Research}
  \bibinfo{volume}{44}, \bibinfo{pages}{939--957}.
\bibitem[{Friedrich et~al.(2020)Friedrich, Krejca, Lagodzinski, Rizzo and
  Zahn}]{friedrich2020memetic}
\bibinfo{author}{Friedrich, T.}, \bibinfo{author}{Krejca, M.S.},
  \bibinfo{author}{Lagodzinski, J.G.}, \bibinfo{author}{Rizzo, M.},
  \bibinfo{author}{Zahn, A.}, \bibinfo{year}{2020}.
\newblock \bibinfo{title}{Memetic genetic algorithms for time series
  compression by piecewise linear approximation}, in:
  \bibinfo{booktitle}{International Conference on Neural Information
  Processing}, pp. \bibinfo{pages}{592--604}.
\bibitem[{Garey and Johnson(1979)}]{garey1979computers}
\bibinfo{author}{Garey, M.}, \bibinfo{author}{Johnson, D.},
  \bibinfo{year}{1979}.
\newblock \bibinfo{title}{Computers and intractability}. volume
  \bibinfo{volume}{174}.
\bibitem[{Grebinski and Kucherov(1998)}]{grebinski1998reconstructing}
\bibinfo{author}{Grebinski, V.}, \bibinfo{author}{Kucherov, G.},
  \bibinfo{year}{1998}.
\newblock \bibinfo{title}{Reconstructing a hamiltonian cycle by querying the
  graph: Application to dna physical mapping}.
\newblock \bibinfo{journal}{Discrete Applied Mathematics} \bibinfo{volume}{88},
  \bibinfo{pages}{147--165}.
\bibitem[{Gutin and Punnen(2006)}]{gutin2006traveling}
\bibinfo{author}{Gutin, G.}, \bibinfo{author}{Punnen, A.P.},
  \bibinfo{year}{2006}.
\newblock \bibinfo{title}{The traveling salesman problem and its variations}.
  volume~\bibinfo{volume}{12}.
\bibitem[{Hamann and Strasser(2016)}]{hamann2016graph}
\bibinfo{author}{Hamann, M.}, \bibinfo{author}{Strasser, B.},
  \bibinfo{year}{2016}.
\newblock \bibinfo{title}{Graph bisection with pareto-optimization}, in:
  \bibinfo{booktitle}{Algorithm Engineering and Experiments}.
\bibitem[{Hao et~al.(2001)Hao, Gui-Lu, Yang and Xiao-Lin}]{hao2001quantum}
\bibinfo{author}{Hao, G.}, \bibinfo{author}{Gui-Lu, L.}, \bibinfo{author}{Yang,
  S.}, \bibinfo{author}{Xiao-Lin, X.}, \bibinfo{year}{2001}.
\newblock \bibinfo{title}{A quantum algorithm for finding a hamilton circuit}.
\newblock \bibinfo{journal}{Communications in Theoretical Physics}
  \bibinfo{volume}{35}, \bibinfo{pages}{385}.
\bibitem[{Harris et~al.(2015)Harris, Berretta, Inostroza{-}Ponta and
  Moscato}]{DBLP:conf/cec/HarrisBIM15}
\bibinfo{author}{Harris, M.}, \bibinfo{author}{Berretta, R.},
  \bibinfo{author}{Inostroza{-}Ponta, M.}, \bibinfo{author}{Moscato, P.},
  \bibinfo{year}{2015}.
\newblock \bibinfo{title}{A memetic algorithm for the quadratic assignment
  problem with parallel local search}, in: \bibinfo{booktitle}{{IEEE} Congress
  on Evolutionary Computation}, pp. \bibinfo{pages}{838--845}.
\bibitem[{Harrow et~al.(2009)Harrow, Hassidim and Lloyd}]{harrow2009quantum}
\bibinfo{author}{Harrow, A.W.}, \bibinfo{author}{Hassidim, A.},
  \bibinfo{author}{Lloyd, S.}, \bibinfo{year}{2009}.
\newblock \bibinfo{title}{Quantum algorithm for linear systems of equations}.
\newblock \bibinfo{journal}{Physical review letters} \bibinfo{volume}{103},
  \bibinfo{pages}{150502}.
\bibitem[{Haythorpe(2015)}]{Haythorpe2015}
\bibinfo{author}{Haythorpe, M.}, \bibinfo{year}{2015}.
\newblock \bibinfo{title}{Fhcp challenge set}.
\newblock \URLprefix \url{http://fhcp.edu.au/fhcpcs}.
\bibitem[{Haythorpe(2019)}]{haythorpe2019fhcp}
\bibinfo{author}{Haythorpe, M.}, \bibinfo{year}{2019}.
\newblock \bibinfo{title}{Fhcp challenge set: The first set of structurally
  difficult instances of the hamiltonian cycle problem}.
\newblock \bibinfo{journal}{arXiv preprint arXiv:1902.10352} .
\bibitem[{Held and Karp(1962)}]{held1962dynamic}
\bibinfo{author}{Held, M.}, \bibinfo{author}{Karp, R.M.}, \bibinfo{year}{1962}.
\newblock \bibinfo{title}{A dynamic programming approach to sequencing
  problems}.
\newblock \bibinfo{journal}{Journal of the Society for Industrial and Applied
  mathematics} \bibinfo{volume}{10}, \bibinfo{pages}{196--210}.
\bibitem[{Helsgaun(2000)}]{HELSGAUN2000106}
\bibinfo{author}{Helsgaun, K.}, \bibinfo{year}{2000}.
\newblock \bibinfo{title}{An effective implementation of the lin–kernighan
  traveling salesman heuristic}.
\newblock \bibinfo{journal}{European Journal of Operational Research}
  \bibinfo{volume}{126}, \bibinfo{pages}{106 -- 130}.
\bibitem[{Holstein and Moscato(1999)}]{10.5555/329055.329079}
\bibinfo{author}{Holstein, D.}, \bibinfo{author}{Moscato, P.},
  \bibinfo{year}{1999}.
\newblock \bibinfo{title}{Memetic algorithms using guided local search: A case
  study}, in: \bibinfo{editor}{Marco~Dorigo, D.C.}, \bibinfo{editor}{Glover,
  F.W.} (Eds.), \bibinfo{booktitle}{New Ideas in Optimization}.
  \bibinfo{publisher}{McGraw-Hill Ltd., UK}, \bibinfo{address}{GBR}, p.
  \bibinfo{pages}{235–244}.
\bibitem[{van Horn et~al.(2018)van Horn, Olij, Sleegers and van~den
  Berg}]{van2018predictive}
\bibinfo{author}{van Horn, G.}, \bibinfo{author}{Olij, R.},
  \bibinfo{author}{Sleegers, J.}, \bibinfo{author}{van~den Berg, D.},
  \bibinfo{year}{2018}.
\newblock \bibinfo{title}{A predictive data analytic for the hardness of
  hamiltonian cycle problem instances}.
\newblock \bibinfo{journal}{Data Analytics} , \bibinfo{pages}{101}.
\bibitem[{Hougardy and Zhong(2021)}]{DBLP:journals/mpc/HougardyZ21}
\bibinfo{author}{Hougardy, S.}, \bibinfo{author}{Zhong, X.},
  \bibinfo{year}{2021}.
\newblock \bibinfo{title}{Hard to solve instances of the euclidean traveling
  salesman problem}.
\newblock \bibinfo{journal}{Math. Program. Computation} \bibinfo{volume}{13},
  \bibinfo{pages}{51--74}.
\bibitem[{Karp(1972)}]{karp1972reducibility}
\bibinfo{author}{Karp, R.}, \bibinfo{year}{1972}.
\newblock \bibinfo{title}{Reducibility among combinatorial problems}, in:
  \bibinfo{booktitle}{Complexity of computer computations}, pp.
  \bibinfo{pages}{85--103}.
\bibitem[{Korte et~al.(2012)Korte, Vygen, Korte and
  Vygen}]{korte2012combinatorial}
\bibinfo{author}{Korte, B.}, \bibinfo{author}{Vygen, J.},
  \bibinfo{author}{Korte, B.}, \bibinfo{author}{Vygen, J.},
  \bibinfo{year}{2012}.
\newblock \bibinfo{title}{Combinatorial optimization}.
  volume~\bibinfo{volume}{5}.
\bibitem[{Lawler(1985)}]{lawler1985traveling}
\bibinfo{author}{Lawler, E.L.}, \bibinfo{year}{1985}.
\newblock \bibinfo{title}{The traveling salesman problem: a guided tour of
  combinatorial optimization}.
\newblock \bibinfo{journal}{Wiley-Interscience Series in Discrete Mathematics}
  .
\bibitem[{Lee et~al.(1999)Lee, Kim, Kim and Sohn}]{lee1999dna}
\bibinfo{author}{Lee, C.M.}, \bibinfo{author}{Kim, S.W.}, \bibinfo{author}{Kim,
  S.M.}, \bibinfo{author}{Sohn, U.}, \bibinfo{year}{1999}.
\newblock \bibinfo{title}{Dna computing the hamiltonian path problem}.
\newblock \bibinfo{journal}{Molecules and cells} \bibinfo{volume}{9},
  \bibinfo{pages}{464}.
\bibitem[{Li and Malik(2017)}]{li2017learning}
\bibinfo{author}{Li, K.}, \bibinfo{author}{Malik, J.}, \bibinfo{year}{2017}.
\newblock \bibinfo{title}{Learning to optimize neural nets}.
\newblock \bibinfo{journal}{arXiv preprint arXiv:1703.00441} .
\bibitem[{Lin and Kernighan(1973)}]{lin1973effective}
\bibinfo{author}{Lin, S.}, \bibinfo{author}{Kernighan, B.},
  \bibinfo{year}{1973}.
\newblock \bibinfo{title}{An effective heuristic algorithm for the
  traveling-salesman problem}.
\newblock \bibinfo{journal}{Operations research} \bibinfo{volume}{21},
  \bibinfo{pages}{498--516}.
\bibitem[{Li{\'s}kiewicz and Schuster(2014)}]{liskiewicz2014new}
\bibinfo{author}{Li{\'s}kiewicz, M.}, \bibinfo{author}{Schuster, M.R.},
  \bibinfo{year}{2014}.
\newblock \bibinfo{title}{A new upper bound for the traveling salesman problem
  in cubic graphs}.
\newblock \bibinfo{journal}{Journal of Discrete Algorithms}
  \bibinfo{volume}{27}, \bibinfo{pages}{1--20}.
\bibitem[{Lucas(2014)}]{lucas2014ising}
\bibinfo{author}{Lucas, A.}, \bibinfo{year}{2014}.
\newblock \bibinfo{title}{Ising formulations of many np problems}.
\newblock \bibinfo{journal}{Frontiers in Physics} \bibinfo{volume}{2},
  \bibinfo{pages}{5}.
\bibitem[{Mahasinghe et~al.(2019)Mahasinghe, Hua, Dinneen and
  Goyal}]{mahasinghe2019solving}
\bibinfo{author}{Mahasinghe, A.}, \bibinfo{author}{Hua, R.},
  \bibinfo{author}{Dinneen, M.J.}, \bibinfo{author}{Goyal, R.},
  \bibinfo{year}{2019}.
\newblock \bibinfo{title}{Solving the hamiltonian cycle problem using a quantum
  computer}, in: \bibinfo{booktitle}{Proceedings of the Australasian Computer
  Science Week Multiconference}, pp. \bibinfo{pages}{1--9}.
\bibitem[{Mahasinghe and Wang(2016)}]{mahasinghe2016efficient}
\bibinfo{author}{Mahasinghe, A.}, \bibinfo{author}{Wang, J.},
  \bibinfo{year}{2016}.
\newblock \bibinfo{title}{Efficient quantum circuits for toeplitz and hankel
  matrices}.
\newblock \bibinfo{journal}{Journal of Physics A: Mathematical and Theoretical}
  \bibinfo{volume}{49}, \bibinfo{pages}{275301}.
\bibitem[{Maretic and Grbic(2015)}]{maretic2015heuristics}
\bibinfo{author}{Maretic, H.P.}, \bibinfo{author}{Grbic, A.},
  \bibinfo{year}{2015}.
\newblock \bibinfo{title}{A heuristics approach to hamiltonian completion
  problem (hcp)}, in: \bibinfo{booktitle}{International Convention on
  Information and Communication Technology, Electronics and Microelectronics},
  pp. \bibinfo{pages}{1607--1612}.
\bibitem[{Mathieson and Moscato(2020)}]{DBLP:conf/ssci/MathiesonM20}
\bibinfo{author}{Mathieson, L.}, \bibinfo{author}{Moscato, P.},
  \bibinfo{year}{2020}.
\newblock \bibinfo{title}{The unexpected virtue of problem reductions or how to
  solve problems being lazy but wise}, in: \bibinfo{booktitle}{Symposium Series
  on Computational Intelligence, {SSCI}}, pp. \bibinfo{pages}{2381--2390}.
\bibitem[{Mendes et~al.(2002)Mendes, Franca and
  Moscato}]{Mendes_fitnesslandscapes}
\bibinfo{author}{Mendes, A.S.}, \bibinfo{author}{Franca, P.M.},
  \bibinfo{author}{Moscato, P.}, \bibinfo{year}{2002}.
\newblock \bibinfo{title}{Fitness landscapes for the total tardiness single
  machine scheduling problem}.
\newblock \bibinfo{journal}{Neural Network World} , \bibinfo{pages}{165--180}.
\bibitem[{Merz and Freisleben(1999)}]{merz1999comparison}
\bibinfo{author}{Merz, P.}, \bibinfo{author}{Freisleben, B.},
  \bibinfo{year}{1999}.
\newblock \bibinfo{title}{A comparison of memetic algorithms, tabu search, and
  ant colonies for the quadratic assignment problem}, in:
  \bibinfo{booktitle}{Proceedings of the 1999 Congress on Evolutionary
  Computation-CEC99 (Cat. No. 99TH8406)}, pp. \bibinfo{pages}{2063--2070}.
\bibitem[{Merz and Freisleben(2002)}]{DBLP:journals/compsys/MerzF02}
\bibinfo{author}{Merz, P.}, \bibinfo{author}{Freisleben, B.},
  \bibinfo{year}{2002}.
\newblock \bibinfo{title}{Memetic algorithms for the traveling salesman
  problem}.
\newblock \bibinfo{journal}{Complex Systems} \bibinfo{volume}{13}.
\bibitem[{Michalewicz(1996)}]{michalewicz1996genetic}
\bibinfo{author}{Michalewicz, Z.}, \bibinfo{year}{1996}.
\newblock \bibinfo{title}{Genetic algorithms+ data structures= evolution
  programs, 3rd edn.{\copyright} springer}.
\bibitem[{Moscato(1993)}]{DBLP:journals/anor/Moscato93}
\bibinfo{author}{Moscato, P.}, \bibinfo{year}{1993}.
\newblock \bibinfo{title}{An introduction to population approaches for
  optimization and hierarchical objective functions: {A} discussion on the role
  of tabu search}.
\newblock \bibinfo{journal}{Annals {OR}} \bibinfo{volume}{41},
  \bibinfo{pages}{85--121}.
\bibitem[{Moscato and Cotta(2019)}]{Moscato2019}
\bibinfo{author}{Moscato, P.}, \bibinfo{author}{Cotta, C.},
  \bibinfo{year}{2019}.
\newblock \bibinfo{title}{An Accelerated Introduction to Memetic Algorithms}.
  \bibinfo{publisher}{Springer International Publishing},
  \bibinfo{address}{Cham}.
\newblock pp. \bibinfo{pages}{275--309}.
\newblock \URLprefix \url{https://doi.org/10.1007/978-3-319-91086-4_9},
  \DOIprefix\doi{10.1007/978-3-319-91086-4_9}.
\bibitem[{Moscato and Mathieson(2019)}]{DBLP:books/sp/19/MoscatoM19}
\bibinfo{author}{Moscato, P.}, \bibinfo{author}{Mathieson, L.},
  \bibinfo{year}{2019}.
\newblock \bibinfo{title}{Memetic algorithms for business analytics and data
  science: {A} brief survey}, in: \bibinfo{editor}{Moscato, P.},
  \bibinfo{editor}{de~Vries, N.J.} (Eds.), \bibinfo{booktitle}{Business and
  Consumer Analytics: New Ideas}, pp. \bibinfo{pages}{545--608}.
\bibitem[{Moscato and Norman(1992)}]{moscato1992memetic}
\bibinfo{author}{Moscato, P.}, \bibinfo{author}{Norman, M.G.},
  \bibinfo{year}{1992}.
\newblock \bibinfo{title}{A memetic approach for the traveling salesman problem
  implementation of a computational ecology for combinatorial optimization on
  message-passing systems}.
\newblock \bibinfo{journal}{Parallel computing and transputer applications}
  \bibinfo{volume}{1}, \bibinfo{pages}{177--186}.
\bibitem[{Moscato and Norman(1998)}]{DBLP:journals/informs/MoscatoN98}
\bibinfo{author}{Moscato, P.}, \bibinfo{author}{Norman, M.G.},
  \bibinfo{year}{1998}.
\newblock \bibinfo{title}{On the performance of heuristics on finite and
  infinite fractal instances of the euclidean traveling salesman problem}.
\newblock \bibinfo{journal}{Journal on Computing} \bibinfo{volume}{10},
  \bibinfo{pages}{121--132}.
\bibitem[{Moscato et~al.(2020)Moscato, Sun and
  Haque}]{DBLP:conf/cec/MoscatoSH20}
\bibinfo{author}{Moscato, P.}, \bibinfo{author}{Sun, H.},
  \bibinfo{author}{Haque, M.N.}, \bibinfo{year}{2020}.
\newblock \bibinfo{title}{Analytic continued fractions for regression: Results
  on 352 datasets from the physical sciences}, in: \bibinfo{booktitle}{{IEEE}
  Congress on Evolutionary Computation, {CEC}}, pp. \bibinfo{pages}{1--8}.
\bibitem[{Moscato and Tinetti(1992)}]{Moscato_Tinetti_92}
\bibinfo{author}{Moscato, P.}, \bibinfo{author}{Tinetti, F.},
  \bibinfo{year}{1992}.
\newblock \bibinfo{title}{Blending heuristics with a population-basedapproach:
  A memetic algorithm for the traveling salesman problem}.
\newblock \bibinfo{type}{report}. Universidad Nacional de La Plata, C.C. 75,
  1900 La Plata, Argentina.
\bibitem[{Mu et~al.(2018)Mu, Dubois{-}Lacoste, Hoos and
  St{\"{u}}tzle}]{DBLP:journals/heuristics/MuDHS18}
\bibinfo{author}{Mu, Z.}, \bibinfo{author}{Dubois{-}Lacoste, J.},
  \bibinfo{author}{Hoos, H.H.}, \bibinfo{author}{St{\"{u}}tzle, T.},
  \bibinfo{year}{2018}.
\newblock \bibinfo{title}{On the empirical scaling of running time for finding
  optimal solutions to the {TSP}}.
\newblock \bibinfo{journal}{Journal of Heuristics} \bibinfo{volume}{24},
  \bibinfo{pages}{879--898}.
\bibitem[{Nazari et~al.(2018)Nazari, Oroojlooy, Snyder and
  Tak{\'a}c}]{nazari2018reinforcement}
\bibinfo{author}{Nazari, M.}, \bibinfo{author}{Oroojlooy, A.},
  \bibinfo{author}{Snyder, L.}, \bibinfo{author}{Tak{\'a}c, M.},
  \bibinfo{year}{2018}.
\newblock \bibinfo{title}{Reinforcement learning for solving the vehicle
  routing problem}, in: \bibinfo{booktitle}{Advances in Neural Information
  Processing Systems}, pp. \bibinfo{pages}{9839--9849}.
\bibitem[{Neri and Zhou(2020)}]{DBLP:conf/cec/NeriZ20}
\bibinfo{author}{Neri, F.}, \bibinfo{author}{Zhou, Y.}, \bibinfo{year}{2020}.
\newblock \bibinfo{title}{Covariance local search for memetic frameworks: {A}
  fitness landscape analysis approach}, in: \bibinfo{booktitle}{{IEEE} Congress
  on Evolutionary Computation, {CEC}}, pp. \bibinfo{pages}{1--8}.
\bibitem[{Oltean(2008)}]{oltean2008solving}
\bibinfo{author}{Oltean, M.}, \bibinfo{year}{2008}.
\newblock \bibinfo{title}{Solving the hamiltonian path problem with a
  light-based computer}.
\newblock \bibinfo{journal}{Natural Computing} \bibinfo{volume}{7},
  \bibinfo{pages}{57--70}.
\bibitem[{Ore(1960)}]{ore1960note}
\bibinfo{author}{Ore, O.}, \bibinfo{year}{1960}.
\newblock \bibinfo{title}{Note on hamilton circuits}.
\newblock \bibinfo{journal}{American Mathematical Monthly}
  \bibinfo{volume}{67}, \bibinfo{pages}{55}.
\bibitem[{Papadimitriou and Steiglitz(1998)}]{papadimitriou1998combinatorial}
\bibinfo{author}{Papadimitriou, C.H.}, \bibinfo{author}{Steiglitz, K.},
  \bibinfo{year}{1998}.
\newblock \bibinfo{title}{Combinatorial optimization: algorithms and
  complexity}.
\bibitem[{Ponta et~al.(2020)Ponta, Dorn, Escobar, {d. L. Correa}, Rosas,
  Hidalgo and Marin}]{INOSTROZAPONTA2020101087}
\bibinfo{author}{Ponta, M.I.}, \bibinfo{author}{Dorn, M.},
  \bibinfo{author}{Escobar, I.}, \bibinfo{author}{{d. L. Correa}, L.},
  \bibinfo{author}{Rosas, E.}, \bibinfo{author}{Hidalgo, N.},
  \bibinfo{author}{Marin, M.}, \bibinfo{year}{2020}.
\newblock \bibinfo{title}{Exploring the high selectivity of 3-d protein
  structures using distributed memetic algorithms}.
\newblock \bibinfo{journal}{Journal of Computational Science}
  \bibinfo{volume}{41}, \bibinfo{pages}{101087}.
\bibitem[{Pulji{\'c} and Manger(2020)}]{puljic2020evolutionary}
\bibinfo{author}{Pulji{\'c}, K.}, \bibinfo{author}{Manger, R.},
  \bibinfo{year}{2020}.
\newblock \bibinfo{title}{Evolutionary operators for the hamiltonian completion
  problem}.
\newblock \bibinfo{journal}{Soft Computing} , \bibinfo{pages}{1--16}.
\bibitem[{Rahman and Kaykobad(2005)}]{rahman2005Hamiltonian}
\bibinfo{author}{Rahman, M.}, \bibinfo{author}{Kaykobad, M.},
  \bibinfo{year}{2005}.
\newblock \bibinfo{title}{On hamiltonian cycles and hamiltonian paths}.
\newblock \bibinfo{journal}{Information Processing Letters}
  \bibinfo{volume}{94}, \bibinfo{pages}{37--41}.
\bibitem[{Ravikumar(1992)}]{ravikumar1992parallel}
\bibinfo{author}{Ravikumar, C.}, \bibinfo{year}{1992}.
\newblock \bibinfo{title}{Parallel techniques for solving large scale
  travelling salesperson problems}.
\newblock \bibinfo{journal}{Microprocessors and Microsystems}
  \bibinfo{volume}{16}, \bibinfo{pages}{149--158}.
\bibitem[{Rubin(1974)}]{rubin1974search}
\bibinfo{author}{Rubin, F.}, \bibinfo{year}{1974}.
\newblock \bibinfo{title}{A search procedure for hamilton paths and circuits}.
\newblock \bibinfo{journal}{Journal of the ACM} \bibinfo{volume}{21},
  \bibinfo{pages}{576--580}.
\bibitem[{Sapin et~al.(2016)Sapin, Jong and Shehu}]{DBLP:conf/gecco/SapinJS16}
\bibinfo{author}{Sapin, E.}, \bibinfo{author}{Jong, K.A.D.},
  \bibinfo{author}{Shehu, A.}, \bibinfo{year}{2016}.
\newblock \bibinfo{title}{A novel ea-based memetic approach for efficiently
  mapping complex fitness landscapes}, in: \bibinfo{editor}{Friedrich, T.},
  \bibinfo{editor}{Neumann, F.}, \bibinfo{editor}{Sutton, A.M.} (Eds.),
  \bibinfo{booktitle}{Genetic and Evolutionary Computation Conference}, pp.
  \bibinfo{pages}{85--92}.
\bibitem[{Seeja(2018)}]{seeja2018hybridham}
\bibinfo{author}{Seeja, K.}, \bibinfo{year}{2018}.
\newblock \bibinfo{title}{Hybridham: A novel hybrid heuristic for finding
  hamiltonian cycle}.
\newblock \bibinfo{journal}{Journal of Optimization} \bibinfo{volume}{2018}.
\bibitem[{Shaikh and Panchal(2012)}]{shaikh2012solving}
\bibinfo{author}{Shaikh, M.}, \bibinfo{author}{Panchal, M.},
  \bibinfo{year}{2012}.
\newblock \bibinfo{title}{Solving asymmetric travelling salesman problem using
  memetic algorithm}.
\newblock \bibinfo{journal}{International Journal of Emerging Technology and
  Advanced Engineering} \bibinfo{volume}{2}, \bibinfo{pages}{634--639}.
\bibitem[{Sleegers and van~den Berg(2020)}]{sleegers2020plant}
\bibinfo{author}{Sleegers, J.}, \bibinfo{author}{van~den Berg, D.},
  \bibinfo{year}{2020}.
\newblock \bibinfo{title}{Plant propagation \& hard hamiltonian graphs}.
\newblock \bibinfo{journal}{Evo} \bibinfo{volume}{2020}, \bibinfo{pages}{10}.
\bibitem[{Sun and Moscato(2019)}]{DBLP:conf/cec/SunM19}
\bibinfo{author}{Sun, H.}, \bibinfo{author}{Moscato, P.}, \bibinfo{year}{2019}.
\newblock \bibinfo{title}{A memetic algorithm for symbolic regression}, in:
  \bibinfo{booktitle}{{IEEE} Congress on Evolutionary Computation, {CEC}}, pp.
  \bibinfo{pages}{2167--2174}.
\bibitem[{Svensson et~al.(2020)Svensson, Tarnawski and
  V{\'e}gh}]{svensson2020constant}
\bibinfo{author}{Svensson, O.}, \bibinfo{author}{Tarnawski, J.},
  \bibinfo{author}{V{\'e}gh, L.A.}, \bibinfo{year}{2020}.
\newblock \bibinfo{title}{A constant-factor approximation algorithm for the
  asymmetric traveling salesman problem}.
\newblock \bibinfo{journal}{Journal of the ACM} \bibinfo{volume}{67},
  \bibinfo{pages}{1--53}.
\bibitem[{Vandegriend and Culberson(1998)}]{vandegriend1998gn}
\bibinfo{author}{Vandegriend, B.}, \bibinfo{author}{Culberson, J.},
  \bibinfo{year}{1998}.
\newblock \bibinfo{title}{The gn, m phase transition is not hard for the
  hamiltonian cycle problem}.
\newblock \bibinfo{journal}{Journal of Artificial Intelligence Research}
  \bibinfo{volume}{9}, \bibinfo{pages}{219--245}.
\bibitem[{Vinyals et~al.(2015)Vinyals, Fortunato and
  Jaitly}]{vinyals2015pointer}
\bibinfo{author}{Vinyals, O.}, \bibinfo{author}{Fortunato, M.},
  \bibinfo{author}{Jaitly, N.}, \bibinfo{year}{2015}.
\newblock \bibinfo{title}{Pointer networks}, in: \bibinfo{booktitle}{Advances
  in Neural Information Processing Systems}, pp. \bibinfo{pages}{2692--2700}.
\bibitem[{Wu et~al.(2018)Wu, Shen and Jiao}]{wu2018game}
\bibinfo{author}{Wu, J.}, \bibinfo{author}{Shen, X.}, \bibinfo{author}{Jiao,
  K.}, \bibinfo{year}{2018}.
\newblock \bibinfo{title}{Game-based memetic algorithm to the vertex cover of
  networks}.
\newblock \bibinfo{journal}{IEEE Transactions on Cybernetics}
  \bibinfo{volume}{49}, \bibinfo{pages}{974--988}.
\bibitem[{Xiao and Nagamochi(2016)}]{xiao2016exact}
\bibinfo{author}{Xiao, M.}, \bibinfo{author}{Nagamochi, H.},
  \bibinfo{year}{2016}.
\newblock \bibinfo{title}{An exact algorithm for tsp in degree-3 graphs via
  circuit procedure and amortization on connectivity structure}.
\newblock \bibinfo{journal}{Algorithmica} \bibinfo{volume}{74},
  \bibinfo{pages}{713--741}.
\bibitem[{Ziobro and Pilipczuk(2019)}]{ziobro2019finding}
\bibinfo{author}{Ziobro, M.}, \bibinfo{author}{Pilipczuk, M.},
  \bibinfo{year}{2019}.
\newblock \bibinfo{title}{Finding hamiltonian cycle in graphs of bounded
  treewidth: Experimental evaluation}.
\newblock \bibinfo{journal}{Journal of Experimental Algorithmics}
  \bibinfo{volume}{24}, \bibinfo{pages}{1--18}.

\end{thebibliography}

\end{document}